\documentclass[lettersize,journal]{IEEEtran}

\usepackage{amsmath,amsfonts}
\usepackage{array}
\usepackage{textcomp}
\usepackage{stfloats}
\usepackage{url}
\usepackage{verbatim}
\usepackage{graphicx}
\usepackage{wrapfig}
\usepackage{booktabs}
\usepackage{bbding}
\usepackage{xcolor}
\usepackage{multirow}
\usepackage{enumitem}
\usepackage[utf8]{inputenc}
\usepackage[T1]{fontenc}
\usepackage{nicefrac}
\usepackage{microtype}
\usepackage{varwidth}
\usepackage{makecell}
\usepackage{colortbl}
\usepackage[most]{tcolorbox}
\usepackage{soul}
\usepackage{adjustbox}
\usepackage{orcidlink}

\usepackage{cite}

\usepackage{subcaption}  

\usepackage{caption}
\usepackage{capt-of}

\usepackage[ruled,vlined]{algorithm2e}

\usepackage{hyperref}

\makeatletter
\@ifundefined{c@lofdepth}{\let\c@lofdepth\relax}{}
\@ifundefined{c@lotdepth}{\let\c@lotdepth\relax}{}
\makeatother

\usepackage{tocloft} 

\definecolor{ForestGreen}{RGB}{34,139,34}

\begin{document}

\title{Never Seen Before: Benchmarking Genuine Zero-Shot Composed Image Retrieval with Consistent Video-Sourced Datasets}

\author{Zhenyu Yang$^{*}$\,\orcidlink{0009-0005-5298-0543}, Zemin Du$^{*}$\,\orcidlink{0009-0007-9385-5452}, Shengsheng Qian\,\orcidlink{0000-0001-9488-2208},~\IEEEmembership{Member,~IEEE,} and Changsheng Xu\,\orcidlink{0000-0001-8343-9665},~\IEEEmembership{Fellow,~IEEE}


\thanks{$^{*}$Zhenyu Yang and Zemin Du contributed equally to this work.}
}

\markboth{Preprint, 2026}
{Yang \MakeLowercase{\textit{et al.}}: Never Seen Before: Benchmarking Genuine Zero-Shot Composed Image Retrieval with Consistent Video-Sourced Datasets}

\IEEEpubid{}

\maketitle

\begin{abstract}
Zero-Shot Composed Image Retrieval (ZS-CIR) aims to retrieve a target image based on a query composed of a reference image and a relative caption without training samples. Existing ZS-CIR datasets often suffer from complete irrelevance between reference and target images due to noisy image sources, and do not achieve a true zero-shot scenario as they use public image datasets that models like CLIP have been trained on. To tackle these challenges, we introduce ZeroSight, a novel benchmark for ZS-CIR. It includes a dataset with consistent reference-target pairs sourced from videos, a data construction pipeline, and evaluation methods that consider the ranking of multiple positive and negative target images. We ensure visually and semantically consistent reference-target pairs by extracting frames from a single video and generating relative captions using LLM-assisted methods. To ensure a true zero-shot scenario, we use video data published after March 31, 2022, ensuring it was not included in CLIP's pre-training data. Additionally, we propose a training-free MLLM-driven method, SC4CIR (Symmetric Consistency for CIR), which can effectively identify hard negative targets through 3 symmetric consistency checks. This method is plug-and-play, seamlessly integrating with various CIR methods and significantly improving performance. Our experimental results from 27 methods reveal that current ZS-CIR datasets and evaluation metrics result in inflated retrieval performance, exaggerating the capabilities of CIR methods. Our benchmark and models can be accessed at \url{https://github.com/sotayang/ZeroSight}.
\end{abstract}


\section{Introduction}

\IEEEPARstart{C}{omposed} Image Retrieval (CIR) \cite{zhang2021heterogeneous, baldrati2022effective, wen2023target, wen2023self} extends traditional cross-modal retrieval tasks by integrating textual descriptions into the image search process, which allows users to search for images based on visual attributes while specifying particular modifications to the query image.
%
%
Traditional CIR methods require labor-intensive creation of annotated triplets consisting of a reference image, modification text, and target image for training.
To mitigate this issue, the recent Zero-Shot CIR (ZS-CIR) task \cite{saito2023pic2word, baldrati2023zero, tang2023context} has been introduced, which aims to enhance the generalization capabilities of CIR methods without relying on annotated triplets, while still maintaining high retrieval accuracy.

Rich datasets~\cite{baldrati2023zero,liu2021image,vaze2023genecis,wu2021fashion, Levy_Ben-Ari_Darshan_Lischinski_2024} for CIR integrate compositional learning~\cite{kim2021dual, hou2020visual, sun2025leveraging, tianccin, ventura2024covr, liu2024bi, xu2024sentence, tan2025dynamic} with image retrieval. 
\begin{figure}[!t]
\centering
\includegraphics[width=\linewidth]{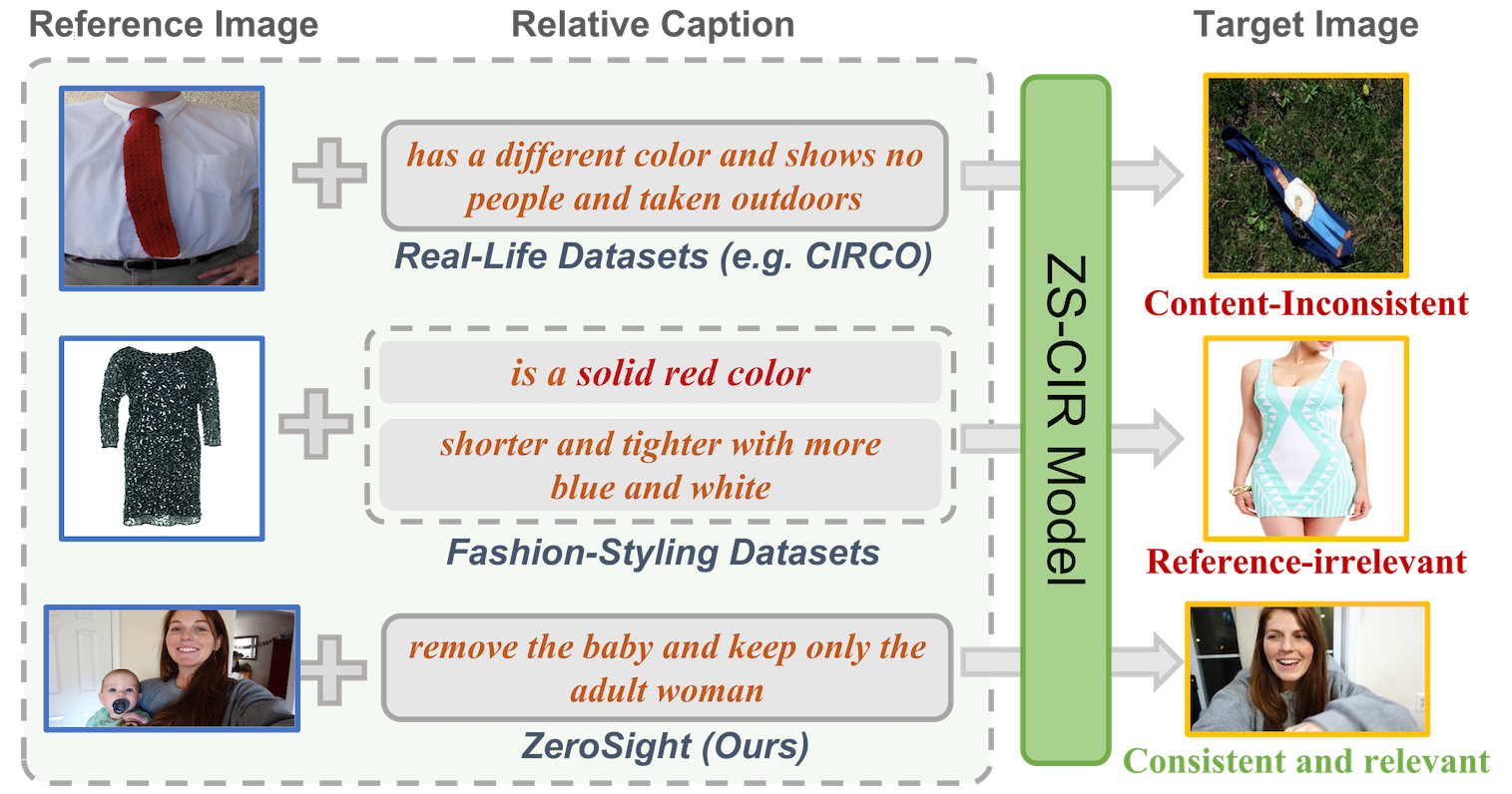}
\caption{\small Comparison of CIR among existing datasets. \rm For real-life datasets, the image pairs are semantically and visually inconsistent, while for fashion datasets, the caption is noisy and the target is irrelevant to the reference (tabulated in Sec.~\ref{categorization}). The image pairs of ZeroSight come from the same video and are both consistent and relevant.}
    \vspace{-7mm}
\label{fig:example}
\end{figure}
This fusion creates a range of challenging tasks that have been widely applied in fashion styling~\cite{wu2021fashion, gao2020fashion} and conditional search~\cite{baldrati2022effective, vaze2023genecis, cao2025multilingual}.
For instance, FashionIQ~\cite{wu2021fashion} focuses on image retrieval in fashion styling, while GeneCIS~\cite{vaze2023genecis} emphasizes the ability of models to adapt to various similarity conditions. However, these datasets typically have weakly related reference images and relatively noisy text descriptions, as shown in Figure~\ref{fig:example}. 
CIRR~\cite{liu2021image} focuses on real-life images, derived from 
NLVR2~\cite{suhr2017corpus}. However, it suffers from two main issues. 
First, the dataset contains several false negatives, which can lead to inaccurate evaluations. 
Second, the queries often do not consider the visual content of the reference image, making the task solvable with standard text-to-image techniques. CIRCO~\cite{baldrati2023zero}, derived from MS COCO~\cite{lin2014microsoft}, addresses these issues by a strategy that leverages CIR methods to ease the annotation process of multiple ground truths.
These datasets provide a diverse set of images, making them valuable resources for advancing research in CIR.

\begin{figure}[!t]
\centering
\includegraphics[width=3.5in]{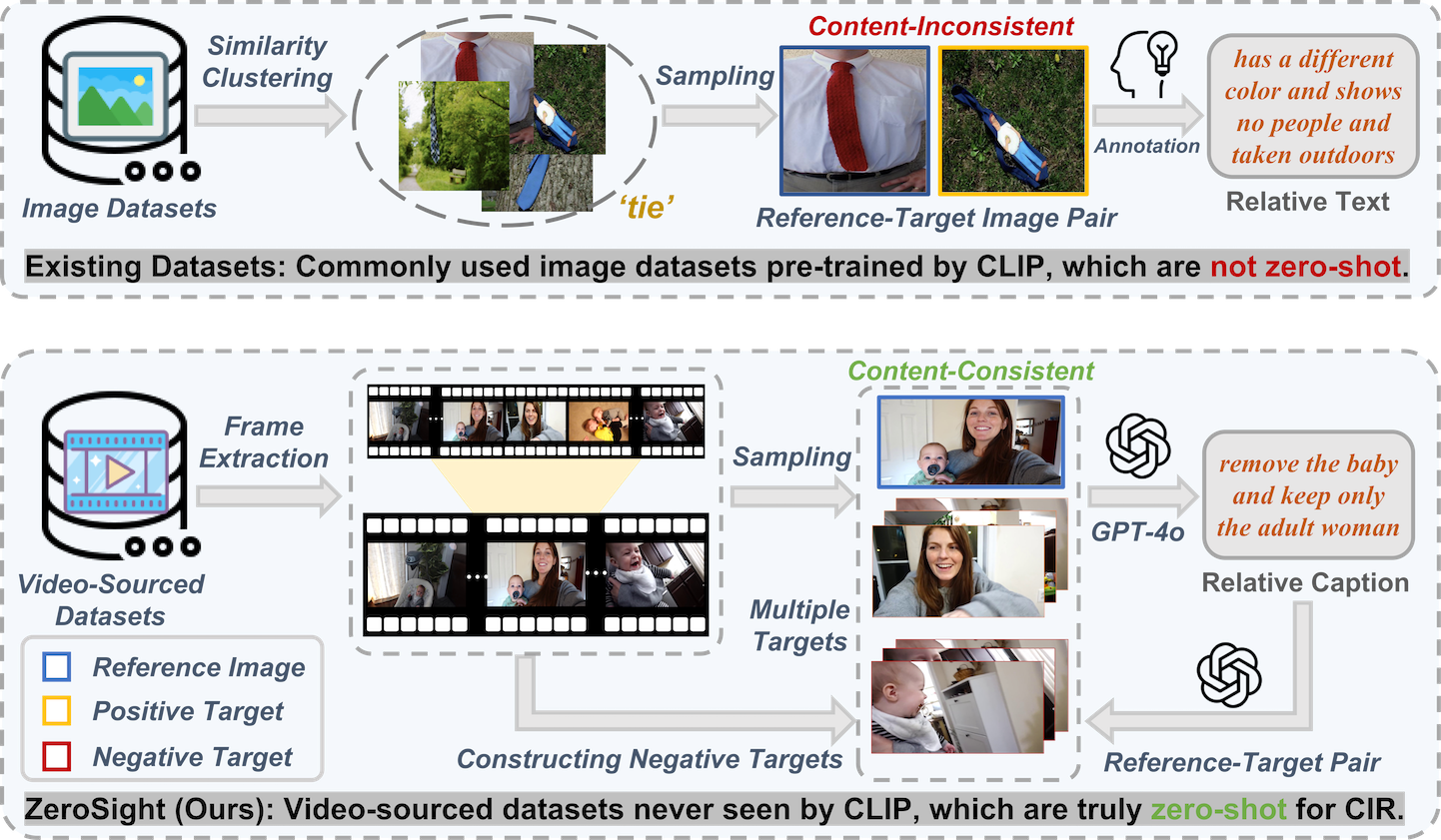}
\caption{\small Comparison of existing ZS-CIR dataset construction pipelines. The current ZS-CIR datasets face two difficulties: (1) There are natural inconsistencies in image pairs constructed from image datasets; (2) The commonly used public image datasets are pre-trained by CLIP. ZeroSight solves both of them by constructing consistent image pairs using video datasets from after 2022.}
\label{fig:comparison}
    \vspace{-5mm}
\end{figure}

However, previous CIR datasets mistakenly treat ZS-CIR as an abstract retrieval task, where the visual and semantic aspects of the target image are not strictly determined by the query. 
The abstract semantic composition causes inconsistencies between reference and target images because these datasets come from noisy image datasets, where there is a natural inconsistency between images of the same abstract concept. 
\IEEEpubidadjcol
For example, CIRR~\cite{liu2021image} constructs similar image sets from NLVR2~\cite{suhr2017corpus} using a clustering-like method and creates reference-target image pairs within these sets. However, NLVR2 is crowdsourced from the web, resulting in gaps even among similar images, which only share abstract conceptual similarities and can be tenuously linked through relative captions. 
For instance, in Figure \ref{fig:comparison}, the content of the image pairs sampled from a cluster is completely unrelated, except for the semantic `tie'. 
In fact, we expect the reference and target images to remain as consistent as possible in aspects not mentioned in the relative caption. From the preceding discussion, we aim to address \textbf{Challenge 1}: How to construct visually and semantically consistent reference-target image pairs in CIR datasets?


Furthermore, most existing datasets overlook a critical point in the zero-shot CIR task: the model should not have access to test image data. However, the image pairs in existing datasets mostly come from commonly used public image datasets such as MS COCO~\cite{lin2014microsoft} and ImageNet~\cite{deng2009imagenet}. 
Additionally, most approaches~\cite{cohen2022my,saito2023pic2word,baldrati2023zero,gu2023language,karthik2023vision,yang2024ldre,yang2024semantic, bao2025mllm, wanggenerative, zhang2025active, chen2025prvr} for ZS-CIR leverage the cross-modal alignment capabilities of large-scale pre-trained Vision-Language Models (VLMs) (e.g., CLIP \cite{radford2021learning}) to extract aligned features from images and text for retrieval, which uses a large number of datasets scraped from the internet, including LAION-2B~\cite{schuhmann2022laion}, DataComp-1B~\cite{gadre2024datacomp}, WIT~\cite{srinivasan2021wit}, WebLI~\cite{chen2022pali}, and DFN-5B~\cite{fang2023data}. 
These datasets cover many commonly used public datasets. For example, CIRCO~\cite{baldrati2023zero} and GeneCIS~\cite{vaze2023genecis} focus on real-life images, constructing image datasets from MS COCO~\cite{lin2014microsoft}, which has been used to train CLIP. Therefore, the existing datasets do not achieve a true zero-shot scenario, inflating results due to the negative impact of pre-trained CLIP (see Sec.~\ref{negative_impact}). Inspired by that, we have to tackle \textbf{Challenge 2}: How to use data that CLIP has never been trained on to construct a truly zero-shot composed image retrieval dataset?

To address these challenges, we introduce \textbf{ZeroSight}, a benchmark with consistent reference-target pairs from video-sourced datasets for ZS-CIR. 
For \textbf{Challenge 1}, we extract frame images from a single video to construct visually and semantically consistent image pairs. We start by collecting high-quality video datasets, then refine and select non-redundant, distinct reference images. For each reference image, we filter and choose multiple content-consistent target images from the same video. Finally, we use LLM-assisted generation to produce relative captions for the consistent image pairs.
For \textbf{Challenge 2}, we utilize video data that CLIP has never been trained on to construct a completely zero-shot CIR dataset. To ensure this, we incorporate video datasets from after March 31, 2022, the release date of the most recent CLIP dataset LAION-2B~\cite{schuhmann2022laion}, ensuring that they have not been included in the pre-training data of CLIP. 

\begin{table*}[t]
\renewcommand{\baselinestretch}{0.9}
\centering
\scriptsize
    %
\caption{\small The comparison of different ZS-CIR datasets. \textbf{\#Index}: the size of the retrieval pool shared by all queries. \textbf{Specific}: whether the dataset is specific to CIR. \textbf{Open-Domain}: whether the query types are diverse. \textbf{Multi-Target}: whether there are multiple ground truths. \textbf{Hard-Negative}: whether to provide hard negative target images. \textbf{Zero-Shot}: whether the images are definitely not pre-trained with CLIP.}
\vspace{-1mm}

\setlength{\tabcolsep}{4mm}{\begin{tabular}{@{}ccccccccc@{}}
\toprule
\textbf{Dataset}                        & \textbf{\#Query} & \textbf{\#Index} & \textbf{Specific}                          & \textbf{Open-Domain}                         & \textbf{Multi-Target}                        & \textbf{Hard-Negative}                       & \textbf{Zero-Shot}                           & \textbf{Image Source}         \\ \midrule
CIRR~\cite{liu2021image}       & 4,148    & 2,316    & \color{ForestGreen}{\CheckmarkBold} & \color{ForestGreen}{\CheckmarkBold} & \color{red}{\XSolidBrush}           & \color{red}{\XSolidBrush}           & \color{red}{\XSolidBrush}           & NLVR2          \\
CIRCO~\cite{baldrati2023zero}  & 800      & 123,403  & \color{ForestGreen}{\CheckmarkBold} & \color{ForestGreen}{\CheckmarkBold} & \color{ForestGreen}{\CheckmarkBold} & \color{red}{\XSolidBrush}           & \color{red}{\XSolidBrush}           & MS COCO        \\
GeneCIS~\cite{vaze2023genecis} & 2,008    & -        & \color{red}{\XSolidBrush}           & \color{ForestGreen}{\CheckmarkBold} & \color{red}{\XSolidBrush}           & \color{red}{\XSolidBrush}           & \color{red}{\XSolidBrush}           & MS COCO        \\
FashionIQ~\cite{wu2021fashion} & 2,005    & 5,179    & \color{ForestGreen}{\CheckmarkBold} & \color{red}{\XSolidBrush}           & \color{red}{\XSolidBrush}           & \color{red}{\XSolidBrush}           & \color{red}{\XSolidBrush}           & Shopping Sites \\ \midrule
\textbf{ZeroSight~(Ours)}               & \textbf{54,740}    & \textbf{197,313}   & \color{ForestGreen}{\CheckmarkBold} & \color{ForestGreen}{\CheckmarkBold} & \color{ForestGreen}{\CheckmarkBold} & \color{ForestGreen}{\CheckmarkBold} & \color{ForestGreen}{\CheckmarkBold} & \textbf{Videos}          \\ \bottomrule
\end{tabular}}

\label{tab:datasets_compare}
\vspace{-0.2cm}
\end{table*}

Additionally, to exclude hard negative targets that are highly similar to the target in retrieval results, we propose a training-free MLLM-driven ZS-CIR method, \textbf{SC4CIR} (Symmetric Consistency for CIR). This method uses a forward retrieval and two reverse processes to identify hard negative targets through consistency checks. It is plug-and-play, seamlessly integrating with various CIR methods and significantly improving performance, especially enhancing LLM-based ZS-CIR methods by 9.12\%.

In summary, our contributions are as follows:
\begin{itemize}
\item
We propose \textbf{ZeroSight}, a novel benchmark with consistent reference-target pairs from video-sourced datasets for ZS-CIR. It encompasses a CIR dataset, a data construction pipeline, and a set of evaluation methods designed to thoroughly assess current CIR methods considering the ranking of multiple positive and negative target images.
\item
We design a multi-stage LLM-assisted dataset construction pipeline to create a truly zero-shot CIR dataset from 12,048 diverse videos, comprising 197,313 candidate images in the retrieval pool and 54,740 queries. On average, each query features 5.16 positive target images and 10.89 negative target images. This is the first dataset to include multiple ground truths and hard negative targets.

\item
We introduce a training-free MLLM-driven method called \textbf{SC4CIR} (Symmetric Consistency for CIR), designed to effectively identify hard negative targets through 3 symmetric consistency checks. This method is plug-and-play, seamlessly integrating with various CIR methods and significantly improving performance.

\item
We release the leaderboard for ZS-CIR on ZeroSight. Extensive experiments conducted on it demonstrate that current CLIP-based CIR methods show inflated results on existing ZS-CIR datasets, exaggerating their capabilities.
\end{itemize}

\section{Related Work}
\subsection{Datasets for Composed Image Retrieval}
Composed Image Retrieval (CIR) task~\cite{vo2019composing, chen2020image, gu2021image, lee2021cosmo, liu2023zero, agnolucci2025isearle, wang2023towards}, focusing on retrieving a target image based on a query of a reference image and a relative caption, has gained significant attention in recent years.
Rich datasets~\cite{liu2021image,vaze2023genecis,wu2021fashion} for CIR integrate compositional learning~\cite{kim2021dual, hou2020visual, sun2025leveraging} with image retrieval, forming a series of demanding tasks, which have been widely applied in fashion styling~\cite{wu2021fashion} and conditional search~\cite{baldrati2022effective, vaze2023genecis, takahashi2014mixture}.
For instance, FashionIQ~\cite{wu2021fashion} focuses on image retrieval in fashion styling, while GeneCIS~\cite{vaze2023genecis} emphasizes the ability of models to adapt to various similarity conditions in conditional search.
The CIRCO~\cite{baldrati2023zero} and CIRR~\cite{liu2021image} datasets, which are derived from MS COCO~\cite{lin2014microsoft} and NLVR2~\cite{suhr2017corpus}, focus on common objects and real-life images. 
%
%
However, they treat ZS-CIR as an abstract retrieval task, leading to inconsistencies between reference and target images. 
%
To overcome it, we extract frame images from high-quality videos to create visually and semantically consistent image pairs.

\subsection{Vision-Language Models for ZS-CIR}
The popularity of the pre-trained BERT~\cite{devlin2018bert} model has sparked interest in developing pre-trained Vision-Language Models (VLMs), including~\cite{chen2020uniter,li2019visualbert,li2020oscar,lu2019vilbert,tan2019lxmert, ramzi2025optimization, li2025attack}, aiming to create Transformer-based~\cite{vaswani2017attention} models trained on large-scale image-text triplets to produce vision-and-language representations.
For CIR, to map images and text into a shared embedding space, many methods harness large pre-trained VLMs, such as CLIP~\cite{radford2021learning}, as the backbone for feature extraction. These models~\cite{baldrati2022conditioned, han2023fame, karthik2023vision} have recently gained popularity due to their exceptional ability to handle multi-modal data. 
%
%
However, most existing ZS-CIR datasets fail to ensure a true zero-shot scenario as they use public image datasets that CLIP have been trained on. To address this, we construct a zero-shot CIR dataset using video data sourced after March 31, 2022, ensuring it has not been included in CLIP's pre-training data.

\subsection{Multi-modal Large Language Models} 
Building on the powerful language capabilities of Large Language Models (LLMs), such as GPT-4 \cite{achiam2023gpt} and LLaMA \cite{touvron2023llama}, recent work has explored integrating multi-modal information, leading to MLLMs~\cite{yangsvbench,yang2026livestar,zhang2026querystream}. Models like GPT-4v, GPT-4o\cite{achiam2023gpt}, and Gemini\cite{team2023gemini} pioneered unified architectures for diverse vision-language tasks beyond simple projection. These models are being applied in an expanding array of fields, where they exhibit outstanding performance. which are pre-trained to integrate visual information into LLM and are post-trained to align with users. OSrCIR\cite{tang2025reason} exemplifies this advancement by leveraging MLLMs to jointly preserve visual semantics and linguistic intent through unified latent space modeling. While prior CIR methods often relied on fine-tuning specialized modules derived from such models, our research demonstrates that effectively combining vision-language models with an LLM enables zero-shot CIR without additional training. Specifically, in this work, we leverage the MLLM's reasoning capability to infer the differences between reference and target images and to generate the reverse instruction based on the multi-modal query.

\section{Problem Statement}
\label{problem_statement}
\textbf{Zero-Shot Composed Image Retrieval (ZS-CIR): } The task of Composed Image Retrieval (CIR) can be defined as a multi-modal retrieval problem. Given a reference image $I^r$ and a relative caption $T$, the objective is to retrieve the target image $I^t$ from an image database $\mathcal{D}$ that aligns with the relative caption while preserving the underlying semantic content of the image that has not been explicitly mentioned. 
Zero-shot CIR further requires that no training samples are available, which thereby should be conducted with out-of-shelf tools.

\section{Dataset: ZeroSight}
\label{dataset}
\begin{figure*}[!t]
\centering
\includegraphics[width=0.95\textwidth]{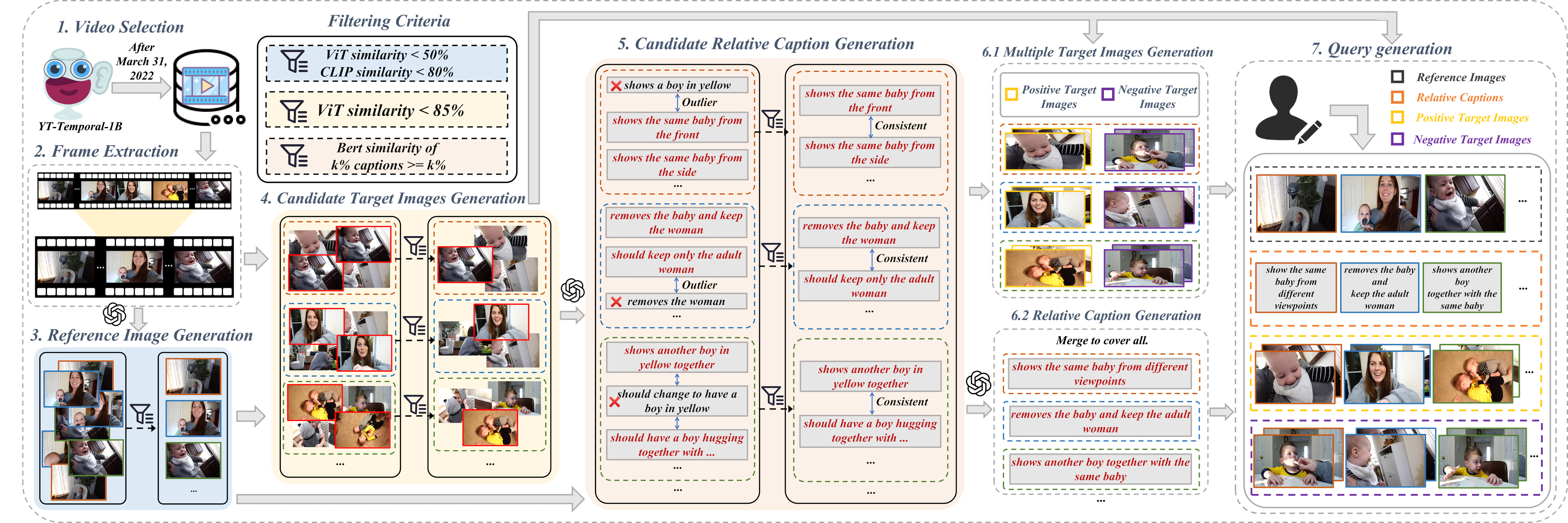}
\captionsetup{width=0.95\textwidth}
\caption{\small \textbf{Overview of the proposed ZeroSight framework.} We design a multi-stage LLM-assisted dataset construction pipeline to create a truly zero-shot CIR dataset from diverse videos, comprising 197,313 candidate images in the retrieval pool and 54,740 queries. This is the first dataset to include multiple ground truths and hard negative target images. In the pipeline, each step is presented \textit{sequentially} using numerical order. And three different screening criteria are involved, each indicated by \textit{a distinct background color}.}
\label{fig:framework}
\end{figure*}

We develop a comprehensive data collection pipeline to construct a high-quality ZS-CIR dataset, tailored for annotating consistent reference-target pairs, as shown in Figure \ref{fig:framework}. The comparison of our proposed ZeroSight with other ZS-CIR datasets is shown in Table \ref{tab:datasets_compare}.

\subsection{Data Collection}
To ensure that our dataset truly aligns with the definition of ZS-CIR, we do not utilize any existing image datasets or collect images from various sources. Instead, we pioneer the use of video datasets as the initial data sources. Then we extract frames from each video to use as images for our dataset. This approach leverages the inherent narrative continuity of videos, ensuring that within a single video, there are pairs of frames that, while similar in content (such as people, objects, etc.), differ in appearance. 
%
To differentiate it from LAION-5B \cite{schuhmann2022laion}, we filter out videos that were published after March 31, 2022 from YT-Temporal-1B \cite{zellers2022merlot}, ensuring that the video data are sufficiently recent. Additionally, to ensure the a rich variety of the images in our dataset, we further categorize the filtered videos. The final videos are divided into 12 main categories and 36 subcategories. Details about the distribution of video categories are included in Sec.~\ref{statistical_ana}.

\subsection{Dataset Construction Pipeline}
\label{construction_pipeline}
We propose a semi-automated dataset construction pipeline for generating multiple ground-truth reference-target image pairs with relative captions to construct a ZS-CIR dataset from videos. This pipeline comprises a multi-stage LLM-assisted generation process and a series of interleaved filtering processes,  
which takes about 3 months and involves over 30 professional annotators. 
Although our dataset is derived from a collection of frames extracted from diverse videos, when constructing the reference-target image pairs, we utilize only the frame images $F = \{f_i\}$ from a single video to ensure the consistency and validity of the constructed results. Detailed prompts are provided in Sec.~\ref{prompt}.

\paragraph{Generating Reference Images}
To ensure the diversity of ZeroSight, the reference images are non-redundant and exhibit significant differences. 
Firstly, we divide video frames into equally spaced subsets $c_i$, each containing 10 frames, resulting in the set of equally spaced subsets $C = \{c_{i} |\ 1 \leq i \leq \lceil \frac{|F|}{10} \rceil \}$. 
Then, we employ MLLMs (such as GPT-4o~\cite{achiam2023gpt}) to initially filter these subsets and select the candidate reference image $r_j$, as follows:
\begin{equation}
r_{j} = 
\begin{cases} 
\text{MLLM}(p_r^1,c_{i}) & \text{if } j = 1 \\
\text{MLLM}(p_r^2,c_{i}, r_{j-1})  & \text{if } j > 1
\end{cases}.
\end{equation}

Thus, we obtain the set of candidate reference images $Rt=\{r_{j} | 1 \leq j \leq |C|\}$. Here, $p_r^1$ and $p_r^2$ are the prompts used to avoid similar images when initially selecting candidate reference images using MLLMs. 
To further ensure the relative independence of each selected image, we design additional steps. First, we use Vision Transformer (ViT)~\cite{dosovitskiy2020image} to remove visually over-consistent reference images, resulting in the set $Rv = \{ Rv_i \in Rt |\  \text{ViT}(Rv_i, \alpha_1) \}$, which consists of candidate reference images that are relatively independent of each other on a visual level.
%
Here, $\alpha_1=0.50$ is the upper limit of the visual similarity threshold for filtering. Then we use the CLIP model to remove semantically over-consistent reference images. The final result is a set of reference images that are relatively independent both visually and semantically, $R = \{ Rs_i \in Rv |\  \text{CLIP}(Rs_i, \beta_1) \}$,
where $\beta_1=0.80$ is the upper limit of the semantic similarity threshold for filtering.

\paragraph{Generating Multiple Target Images}
After generating the set of reference images from a video, we construct multiple target images from the same video $F$ for each reference image. 
The goal is to select images that have a certain degree of similarity to the reference image but are not completely identical. To achieve this, we first use ViT to filter out $Sv_i = \{ Sv_j^i \in F |\ \text{ViT}(Sv_j^i , Rs_i, \alpha_2, \alpha_3), Rs_i \in R\}$, the set of images that have a certain visual similarity to the $i$-th reference image $Rs_i$, 
where $\alpha_2=0.35$ and $\alpha_3=0.50$ are the lower and upper limits of visual similarity thresholds, respectively. Next, to further ensure a semantic connection between the filtered similar images and the reference image, we consider the semantic similarity of these images to the reference image, and use CLIP to further filter out $Ss_i = \{ Ss_j^i \in Sv |\  \text{CLIP}(Ss_j^i , Rs_i, \beta_2, \beta_3), Rs_i \in R\}$, the set of images with a certain degree of semantic similarity to $Rs_i$, 
where $\beta_2=0.65$ and $\beta_3=0.75$ are respectively the lower and upper limits of semantic similarity thresholds. Finally, to ensure that the set of target images corresponding to each reference image does not contain visually redundant images, we use ViT again to filter out images in $Ss_i$ that are visually over-similar, as shown below:
\begin{equation}
St_i = \{ St_j^i \in Ss_i |\  \text{ViT}(St_j^i, Ss_m^i, \theta), Ss_m^i \in Ss_i, m \neq j \},
\end{equation}
which represents the final set of candidate target images. Here $\theta=0.85$ is the upper limit of the visual similarity threshold for filtering. Next, for the videos that are not used to generate $St=\{St_{i} | 1 \leq i \leq |R|\}$, we extract frame images at equal intervals of every 5 frames. These images are then combined with each $St$ to form the retrieval pool.

\paragraph{Generating Relative Captions}
After generating reference image sets and candidate target image sets, we use MLLMs to generate candidate relative caption for each reference image and each candidate target image in its corresponding candidate set. Then we construct $Tt_i = \{ t_j^i = \text{MLLMs}(Rs_i,St^j) |\ Rs_i \in R , St^j \in St_i \}$, a set of candidate relative captions for the $i$-th reference image $Rs_i$, 
where $St_i$ is the set of candidate target images corresponding to the $i$-th reference image $Rs_i$, $St^j$ is the $j$-th candidate target image in $St^i$, and $t_j^i$ represents the candidate relative caption generated for the $i$-th reference image and the $j$-th candidate target image in $St_i$. Subsequently, we use BERT~\cite{devlin2018bert} to calculate the textual similarity for all pairs of candidate relative captions within $Tt_i$, resulting in the set $Vx = \{ v_x^y = \text{BERT}(x,y) |\ x, y \in Tt_i , y \neq x \}$, which represents the textual similarities calculated between the $x$-th candidate relative caption in $Tt_i$ and all other candidate relative captions in the same set. 
To obtain the final relative caption, we define the following function:
\begin{equation}
k(x) = max\{ \epsilon |\ \frac{|\{v_x^y \in Vx |\ v_x^y \geq \epsilon\}|}{|Vx|} \geq \epsilon\},
\end{equation}
where $k\ (0 \leq k \leq 1)$ is a variable parameter. We then construct the set $T_i$ as follows:
\begin{equation}
Ts_i = \{ t_j \in Tt_i | \text{BERT}(\mathop{\arg\max}\limits_{x \in Tt_i}k(x),t_j) \geq \mathop{\max}\limits_{x \in Tt_i}k(x) \},
\end{equation}
which represents the subset of $Tt_i$ used to generate the final relative caption. Then, we can generate the final target image set $Sp_i$ for the $i$-th reference image $Rs_i$ as follows:
\begin{equation}
Sp_i = \{p_m = St_j |\ St_j \in St , t_j^i \in Tt_i , t_j^i \in Ts_i \},
\end{equation}
which is the set of positive target images for the $i$-th reference image $Rs_i$. Conversely, the set of negative target images for the $i$-th reference image $Rs_i$ is is represented as $Sn_i = \{ n_m \in St |\ n_m \notin Sp_i \}$. This unique characteristic distinguishes our dataset from others.
Subsequently, we utilize LLMs (such as GPT-4) to generate a unified final relative caption $Tl_i = \text{LLM}(Ts_i,p_t)$ for $Rs_i$, along with the corresponding $Sp_i$ and $Sn_i$, 
where $p_t$ represents the prompt used to generate a text that summarizes the meaning of $t_j(t_j \in Ts_j)$. Then we obtain a triplet $(Rs_i, Tl_i, \langle Sp_i, Sn_i \rangle)$ for $Rs_i(Rs_i \in R)$ as the query structure. Finally, to ensure the quality of ZeroSight and MLLM annotations, we manually review and select appropriate queries. Additionally, we utilized three different LVLMs to construct queries following the same pipeline initially. The constructed queries were then manually assessed from four different aspects, as shown in Table \ref{tab:lvlms}. Based on the assessment, we selected GPT-4o for our pipeline.

\begin{figure*}[!t]
\centering

\begin{minipage}[b]{0.48\textwidth}
\centering
\includegraphics[width=\textwidth]{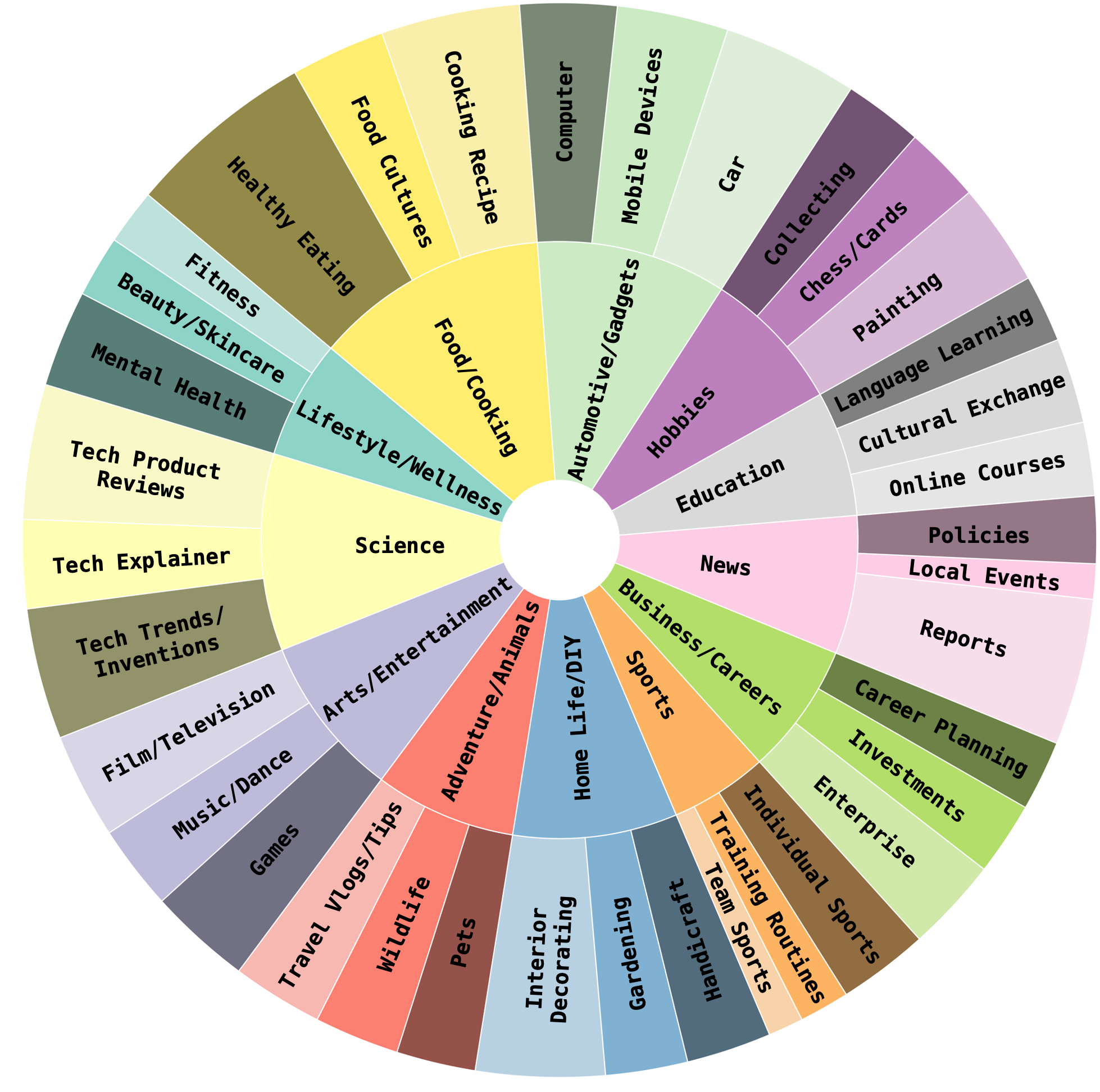}
\captionsetup{width=0.9\textwidth}
\caption{Distribution of videos across 12 primary categories and 36 subcategories, highlighting the diversity and inclusiveness of video types in our dataset.}
\label{fig:video_dis}
\end{minipage}
\hfill 
\begin{minipage}[b]{0.48\textwidth}
\centering
\includegraphics[width=\textwidth]{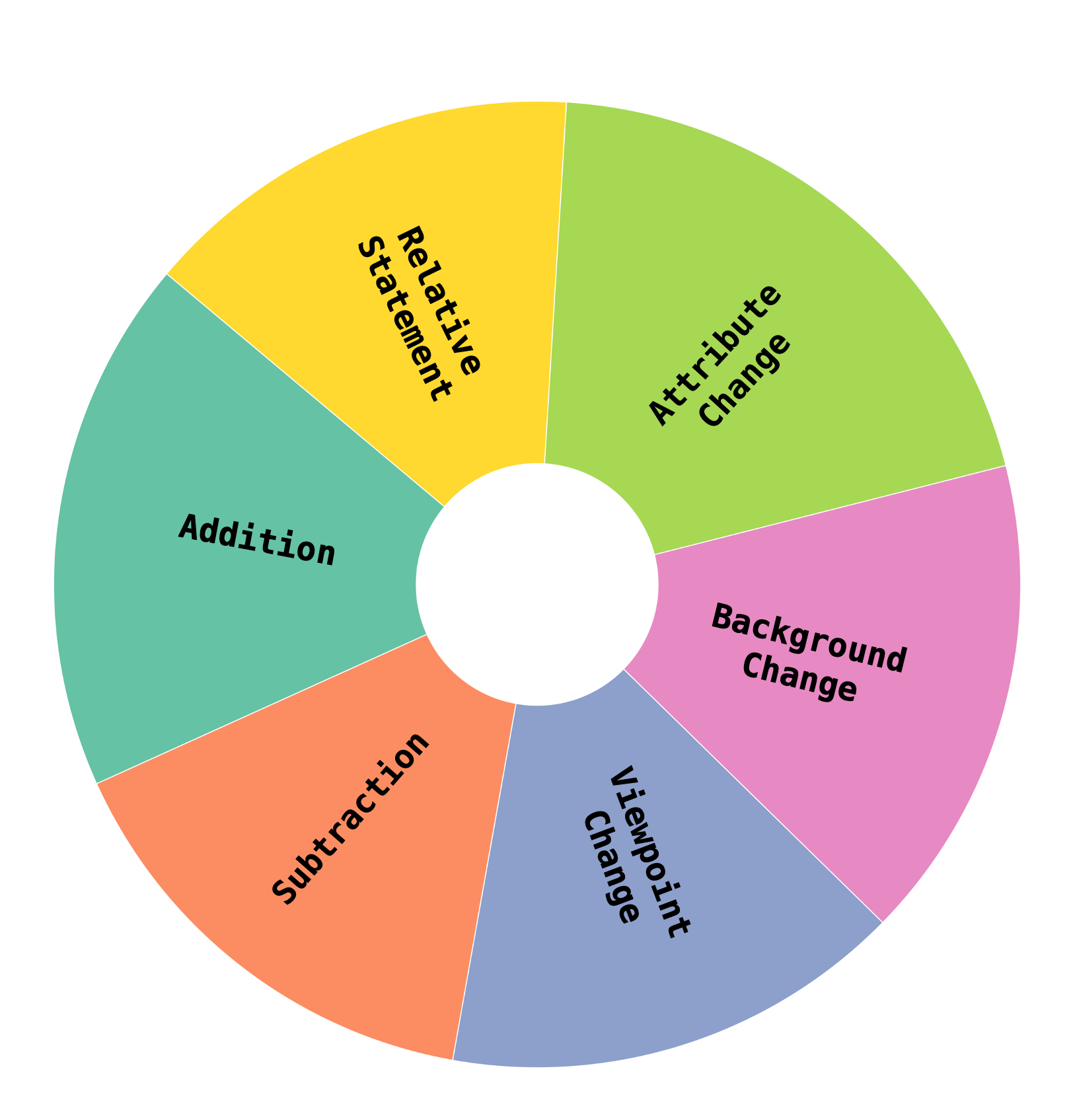}
\captionsetup{width=0.9\textwidth}
\caption{Distribution of queries across 6 distinct categories, each designed to assess a specific aspect of ZS-CIR methods.}
\label{fig:query_dis}
\end{minipage}

\end{figure*}

\subsection{Statistical Analysis}
\label{statistical_ana}
\paragraph{Video Categories}
Our dataset is collected from videos organized into 12 primary categories and 36 subcategories. These categories are depicted in Figure \ref{fig:video_dis}, illustrating the the diversity and inclusiveness of video types within our dataset.

\paragraph{Query Categories}
To facilitate a more comprehensive evaluation for ZS-CIR methods, we classify the
queries into 6 distinct categories as shown in Figure \ref{fig:query_dis}. Each category corresponds to one specific assessment for ZS-CIR methods. The criteria for these categories are as follows: (1) \textit{Addition}: Evaluating retrieval capability when adding elements. (2) \textit{Subtraction}: Evaluating retrieval capability when removing elements. (3) \textit{Background Change}: Evaluating retrieval capability when changing different backgrounds. (4) \textit{Viewpoint Change}: Evaluating retrieval capability when shifting viewpoints. (5) \textit{Attribute Change}: Evaluating retrieval capability when changing attributes of objects. (6) \textit{Relative Statement}: Evaluating retrieval capability involving relative description of objects.

\begin{table}[t]
\renewcommand\arraystretch{0.8}
  \vspace{-1mm}
  \centering
  \small
  \caption{\small Evaluation of LVLMs in constructing queries. \textbf{IEM}: Image Element Moderation. \textbf{MTIA}: Multiple Target Image Abundance. \textbf{TIC}: Text-Image Consistency. \textbf{SC}: Semantic Conciseness.}
  %
\vspace{1mm}
\begin{tabular}{@{}cccccc@{}}
\toprule
\textbf{LVLM}                                       & \textbf{IEM} & \textbf{MTIA} & \textbf{TIC} & \textbf{SC} & \textbf{Average} \\ \midrule
Gemini 1.5 Pro\fontsize{6pt}{\baselineskip}\selectfont     & 3.80        & 3.31         & 4.48         & 4.56         & 4.04         \\
Qwen2.5-VL-72B\fontsize{6pt}{\baselineskip}\selectfont     & 3.62        & 2.64         & 4.29         & 4.52         & 3.78         \\
GPT-4o\fontsize{6pt}{\baselineskip}\selectfont             & \textbf{4.17}        & \textbf{4.44}         & \textbf{4.64}         & \textbf{4.59}         & \textbf{4.46}         \\ 
\bottomrule
\end{tabular}

  \label{tab:lvlms}
   \vspace{-5mm}
\end{table}


\section{Method: SC4CIR}
\label{method}


As mentioned above, existing datasets and methods do not account for hard negatives that are highly similar to the target. In our ZeroSight benchmark, each query includes an average of 10.89 hard negative images, increasing the retrieval difficulty. To address this, we propose a training-free MLLM-driven method, SC4CIR (Symmetric Consistency for CIR), which uses a forward retrieval and two reverse processes to identify hard negative targets through consistency checks. This method is plug-and-play, seamlessly integrating with various CIR methods and significantly improving performance, especially LLM-based CIR methods.
\subsection{Symmetric Consistency Checking}
\begin{figure}[!t]
\centering
\vspace{0.2cm}
\includegraphics[width=3.5in]{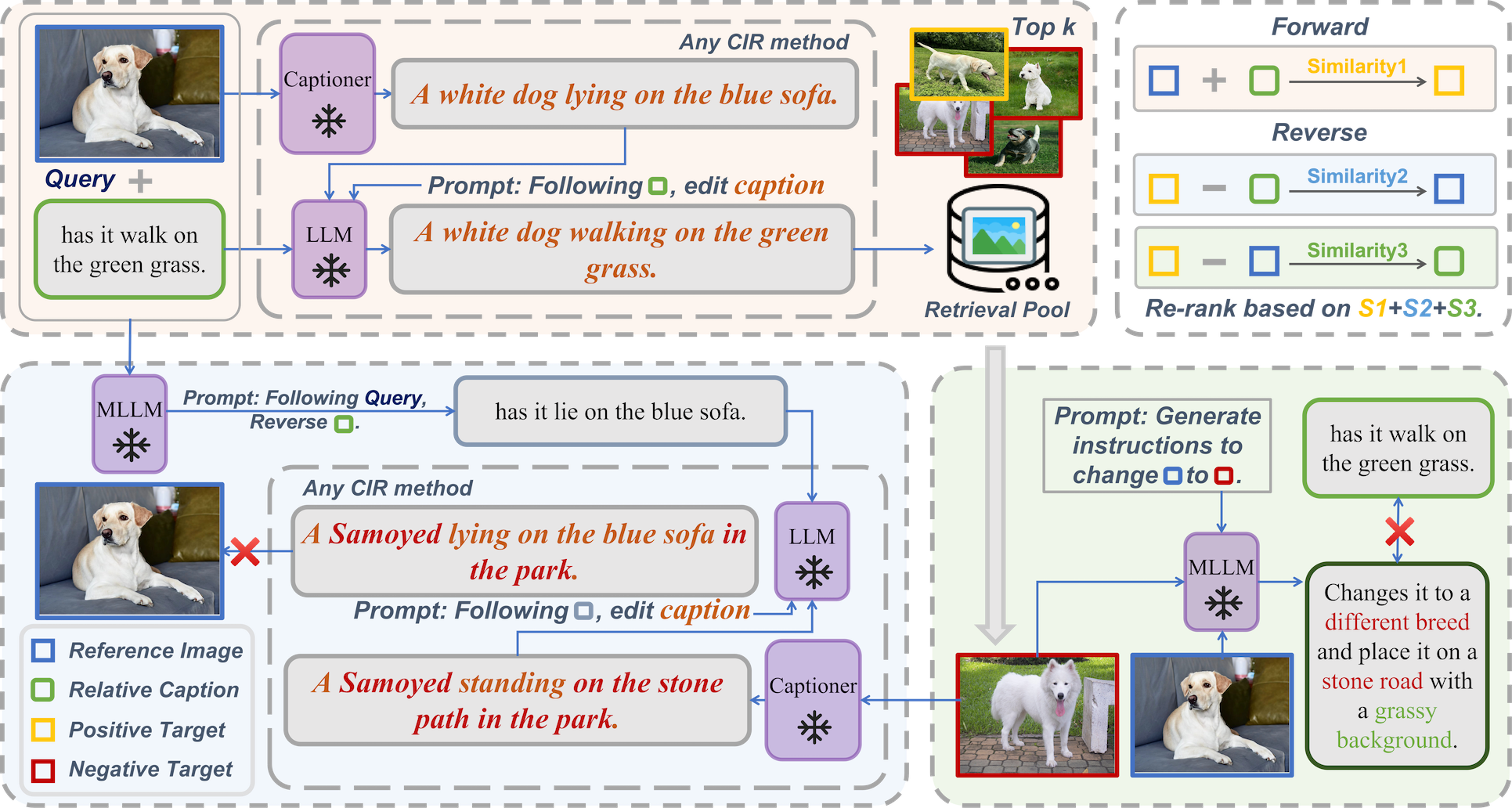}
\caption{\textbf{\small Architecture of the proposed SC4CIR method.} Forward and bidirectional reverse retrieval flows are distinguished by color blocks.}
\label{fig:method}
    \vspace{-5mm}
\end{figure}

As shown in Figure \ref{fig:method}, most CIR methods mistakenly treat hard negative images as targets due to the ambiguity in the forward retrieval process. For example, in the forward process $I^r + T \rightarrow I^t$, where $I^r$ represents the reference image, $T$ represents the relative text and $I^t$ represents the retrieved candidate targets (which may be positive or negative images), CIReVL~\cite{karthik2024visionbylanguage} successfully retrieves images that match the `white dog' semantics. However, most of these images (e.g., `Samoyed') do not match the reference image. Therefore, we introduce two symmetric reverse processes to verify the consistency of the results.
For the top $N$ ($N=30$) candidates retrieved in the forward process, we separately calculate the two similarities in the reverse process as follows:

\paragraph{Reverse Process 1:} $I^t - T \rightarrow I^r$.
%
We replace the subtraction operation with the generation of an instruction $T'$, which is the inverse of $T$, i.e., $I^t + T' \rightarrow I^r$. Since the reverse operation requires information from the reference image, we utilize an MLLM (GPT-4o) to generate $T' = \phi(I^r, T, p_1)$. Here, $\phi$ represents the MLLM used and $p_1$ is a prompt that directs the model to generate the reverse operation $T'$ based on $I^r$ and $T$. We then apply the same CIR method, using $I^t$ and $T'$ as inputs for retrieval. Finally, we calculate the similarity $S_2$ between retrieval results and the reference image $I^r$.


\paragraph{Reverse Process 2:} $I^t - I^r\rightarrow T$.
We achieve the subtraction operation by describing the differences between the reference image $I^r$ and the candidate target $I^t$. We use GPT-4o to generate a relative caption for the candidate target, $T'' = \phi(I^r, I^t, p_2)$, where $p_2$ is a prompt that guides the model to generate the instruction $T''$ that transforms $I^r$ into $I^t$. We then verify the consistency between the original relative caption $T$ and the generated relative caption $T''$ by calculating their similarity $S_3$.

\subsection{Positive-Negative Re-Ranking}
For the top $N$ candidate targets in the retrieval results, the lower the similarity in the reverse process, the worse the symmetric consistency, and the more likely it is to be a negative target. In the forward retrieval process, the similarity with $I^r$ is denoted as $S_1$, and in the reverse processes 1 and 2, the similarities obtained are $S_2$ and $S_3$, respectively. Based on these similarities, we calculate the overall similarity $S = S_1 + S_2 + S_3$ and re-rank the retrieval results accordingly to obtain the final retrieval results.
This result is based on symmetric consistency checking, ensuring the reliability and accuracy. 

\section{Experiments}
\label{experiments}
\subsection{Experimental Setup}
{\label{ex_setup}}
To effectively evaluate the performance of current both ZS-CIR and CIR methods, we meticulously select a diverse range of state-of-the-art methods, as detailed in Sec.~\ref{baseline}. For the two categories of methods, we design two distinct experimental setups within ZeroSight to rigorously assess the capabilities.

\textbf{Evaluation for ZS-CIR Methods.} In this setup, we evaluate the capabilities of ZS-CIR methods. Except for Text-only and Image-only, all other methods are provided with a reference image and a corresponding relative caption. This evaluation represents the core functionality offered by ZeroSight as a ZS-CIR dataset.

\textbf{Evaluation for CIR Methods.} We also provide an evaluation for CIR methods. We divide ZeroSight into training and test sets in a ratio of 9:1, with the two sets containing 49,266 and 5,474 queries respectively. Since the queries in ZeroSight have multiple ground truths, it is essential to create one training data for each reference image, its relative caption and each positive target image. During inference, CIR methods only need to be provided with the reference image and its relative caption. This evaluation broadens the applicability of ZeroSight, making it suitable for both ZS-CIR and CIR, thereby offering a richer selection for different CIR methods.

\begin{table*}[t]
\vspace{-4mm}
\renewcommand\arraystretch{0.60}
\centering
\footnotesize
\caption{Results of comparison among different ZS-CIR methods on ZeroSight. 
}
\vspace{-1mm}

\begin{tabular}{ccccccccccccc}
\toprule
\multirow{3}{*}{\textbf{Backbone}} & \multirow{3}{*}{\textbf{Method}} & \multirow{3}{*}{\textbf{Training-free}} & \multicolumn{10}{c}{\textbf{Evaluation Metrics}}                                                                                                                                 \\ \cmidrule(l){4-13} 
                                   &                                  &                                         & \multicolumn{5}{c|}{\textbf{mAP@k}}                                                               & \multicolumn{5}{c}{\textbf{PNR-mAP@k}}                                       \\
                                   &                                  &                                         & \textbf{k=5} & \textbf{k=10} & \textbf{k=25} & \textbf{k=50} & \multicolumn{1}{c|}{\textbf{Avg.}} & \textbf{k=5} & \textbf{k=10} & \textbf{k=25} & \textbf{k=50} & \textbf{Avg.} \\ \midrule
\multirow{6}{*}{ViT-B/32}          & CIReVL\fontsize{6pt}{\baselineskip}\selectfont{(ICLR'2024)}                & \color{ForestGreen}{\CheckmarkBold}     & 7.76         & 8.51          & 9.57          & 10.12         & \multicolumn{1}{c|}{8.99}          & 4.95         & 6.00          & 7.28          & 7.96          & 6.55          \\
                                   & LDRE\fontsize{6pt}{\baselineskip}\selectfont{(SIGIR’2024)}                 & \color{ForestGreen}{\CheckmarkBold}     & 9.53         & 10.46         & 11.79         & 12.47         & \multicolumn{1}{c|}{11.06}         & 6.01         & 7.33          & 8.93          & 9.77          & 8.01          \\
                                   & SEIZE\fontsize{6pt}{\baselineskip}\selectfont{(MM’2024)}                   & \color{ForestGreen}{\CheckmarkBold}     & 10.64        & 11.61         & 12.94         & 13.65         & \multicolumn{1}{c|}{12.21}         & 6.39         & 7.82          & 9.44          & 10.30         & 8.49          \\
                                   & PALAVRA\fontsize{6pt}{\baselineskip}\selectfont{(ECCV'2022)}               & \color{red}{\XSolidBrush}               & 12.67        & 13.89         & 15.51         & 16.36         & \multicolumn{1}{c|}{14.61}         & 7.99         & 9.70          & 11.62         & 12.61         & 10.48         \\
                                   & SEARLE-OTI\fontsize{6pt}{\baselineskip}\selectfont{(ICCV'2023)}            & \color{red}{\XSolidBrush}               & 21.39        & 24.17         & 26.45         & 27.27         & \multicolumn{1}{c|}{24.82}         & 19.01        & 22.40         & 25.19         & 26.19         & 23.20         \\
                                   & SEARLE\fontsize{6pt}{\baselineskip}\selectfont{(ICCV'2023)}                & \color{red}{\XSolidBrush}               & 22.65        & 25.58         & 27.98         & 28.84         & \multicolumn{1}{c|}{26.26}         & 20.15        & 23.72         & 26.65         & 27.70         & 24.56         \\
                                   \rowcolor{gray!20} & SEIZE+SC4CIR\fontsize{6pt}{\baselineskip}\selectfont{(Ours)}                & \color{ForestGreen}{\CheckmarkBold}               & 12.59        & 13.71         & 15.21         & 16.12         & \multicolumn{1}{c|}{14.41}         & 9.48        & 10.93         & 11.98         & 13.41         & 11.45         \\ \midrule
\multirow{10}{*}{ViT-L/14}         & Captioning                       & \color{ForestGreen}{\CheckmarkBold}     & 5.86         & 6.68          & 7.74          & 8.32          & \multicolumn{1}{c|}{7.15}          & 3.84         & 4.94          & 6.22          & 6.94          & 5.49          \\
                                   & Text-only                        & \color{ForestGreen}{\CheckmarkBold}     & 8.38         & 9.19          & 10.27         & 10.84         & \multicolumn{1}{c|}{9.67}          & 5.26         & 6.40          & 7.68          & 8.34          & 6.92          \\
                                   & CIReVL\fontsize{6pt}{\baselineskip}\selectfont{(ICLR'2024)}                & \color{ForestGreen}{\CheckmarkBold}     & 9.45         & 10.40         & 11.74         & 12.42         & \multicolumn{1}{c|}{11.00}         & 5.95         & 7.29          & 8.91          & 9.74          & 7.97          \\
                                   & Image-only                       & \color{ForestGreen}{\CheckmarkBold}     & 15.62        & 17.26         & 19.38         & 20.54         & \multicolumn{1}{c|}{18.20}         & 5.58         & 7.78          & 10.34         & 11.78         & 8.87          \\
                                   & LDRE\fontsize{6pt}{\baselineskip}\selectfont{(SIGIR’2024)}                 & \color{ForestGreen}{\CheckmarkBold}     & 12.60        & 14.00         & 15.62         & 16.44         & \multicolumn{1}{c|}{14.67}         & 8.01         & 9.89          & 11.87         & 12.87         & 10.66         \\
                                   & SEIZE\fontsize{6pt}{\baselineskip}\selectfont{(MM’2024)}                   & \color{ForestGreen}{\CheckmarkBold}     & 13.02        & 14.36         & 15.98         & 16.80         & \multicolumn{1}{c|}{15.04}         & 8.21         & 10.04         & 12.02         & 13.01         & 10.82         \\
                                   & Pic2Word\fontsize{6pt}{\baselineskip}\selectfont{(CVPR'2023)}              & \color{red}{\XSolidBrush}               & 15.06        & 18.09         & 19.78         & 20.21         & \multicolumn{1}{c|}{18.28}         & 12.63        & 16.32         & 18.46         & 18.96         & 16.59         \\
                                   & SEARLE-OTI\fontsize{6pt}{\baselineskip}\selectfont{(ICCV'2023)}            & \color{red}{\XSolidBrush}               & 24.93        & 30.68         & 33.89         & 34.71         & \multicolumn{1}{c|}{31.05}         & 20.32        & 27.33         & 31.39         & 32.33         & 27.84         \\
                                   & SEARLE\fontsize{6pt}{\baselineskip}\selectfont{(ICCV'2023)}                & \color{red}{\XSolidBrush}               & 26.12        & 32.17         & 35.55         & 36.41         & \multicolumn{1}{c|}{32.56}         & 21.26        & 28.64         & 32.92         & 33.91         & 29.18         \\
                                   & LinCIR\fontsize{6pt}{\baselineskip}\selectfont{(CVPR'2024)}                & \color{red}{\XSolidBrush}               & 28.13        & 34.48         & 38.03         & 38.93         & \multicolumn{1}{c|}{34.89}         & 23.02        & 30.77         & 35.27         & 36.31         & 31.34         \\
                                   \rowcolor{gray!20} & SEIZE+SC4CIR\fontsize{6pt}{\baselineskip}\selectfont{(Ours)}                & \color{ForestGreen}{\CheckmarkBold}               & 15.13        & 16.56         & 18.03         & 18.91         & \multicolumn{1}{c|}{17.16}         & 11.41        & 13.28         & 15.16         & 16.13         & 14.00         \\ 
                                   \rowcolor{gray!20} & LinCIR+SC4CIR\fontsize{6pt}{\baselineskip}\selectfont{(Ours)}                & \color{red}{\XSolidBrush}               & 29.24        & 35.63         & 39.36         & 36.01         & \multicolumn{1}{c|}{35.06}         & 25.13        & 32.96         & 37.35         & 38.50         & 33.49         \\ \midrule
\multirow{6}{*}{ViT-G/14}          & CIReVL\fontsize{6pt}{\baselineskip}\selectfont{(ICLR'2024)}                & \color{ForestGreen}{\CheckmarkBold}     & 12.00        & 13.24         & 14.74         & 15.49         & \multicolumn{1}{c|}{13.87}         & 7.56         & 9.25          & 11.08         & 12.00         & 9.97          \\
                                   & LDRE\fontsize{6pt}{\baselineskip}\selectfont{(SIGIR’2024)}                 & \color{ForestGreen}{\CheckmarkBold}     & 15.98        & 17.64         & 19.64         & 20.65         & \multicolumn{1}{c|}{18.48}         & 10.04        & 12.30         & 14.75         & 15.97         & 13.27         \\
                                   & SEIZE\fontsize{6pt}{\baselineskip}\selectfont{(MM’2024)}                   & \color{ForestGreen}{\CheckmarkBold}     & 16.63        & 18.37         & 20.47         & 21.54         & \multicolumn{1}{c|}{19.25}         & 10.37        & 12.75         & 15.33         & 16.61         & 13.77         \\
                                   & Pic2Word\fontsize{6pt}{\baselineskip}\selectfont{(CVPR'2023)}              & \color{red}{\XSolidBrush}               & 16.07        & 19.39         & 21.25         & 21.73         & \multicolumn{1}{c|}{19.61}         & 13.39        & 17.45         & 19.81         & 20.35         & 17.75         \\
                                   & SEARLE\fontsize{6pt}{\baselineskip}\selectfont{(ICCV'2023)}                & \color{red}{\XSolidBrush}               & 31.46        & 37.95         & 41.72         & 41.83         & \multicolumn{1}{c|}{38.23}         & 23.14        & 31.73         & 35.30         & 36.58         & 31.69         \\
                                   & LinCIR\fontsize{6pt}{\baselineskip}\selectfont{(CVPR'2024)}                & \color{red}{\XSolidBrush}               & 32.00        & 38.57         & 42.38         & 42.51         & \multicolumn{1}{c|}{38.87}         & 24.62        & 33.31         & 37.93         & 38.23         & 33.52         \\
                                   \rowcolor{gray!20} & SEIZE+SC4CIR\fontsize{6pt}{\baselineskip}\selectfont{(Ours)}                & \color{ForestGreen}{\CheckmarkBold}               & 18.11        & 19.76         & 21.93         & 22.98         & \multicolumn{1}{c|}{20.70}         & 12.68        & 15.14         & 17.69         & 18.98         & 16.12         \\
                                   \rowcolor{gray!20} & LinCIR+SC4CIR\fontsize{6pt}{\baselineskip}\selectfont{(Ours)}                & \color{red}{\XSolidBrush}               & 33.04        & 39.68         & 43.57         & 43.79         & \multicolumn{1}{c|}{40.02}         & 26.55        & 35.28         & 39.90         & 40.21         & 35.49         \\ \bottomrule
\end{tabular}
\label{tab:ZS res}
\vspace{-4mm}
\end{table*}

\subsection{Baselines}
\label{baseline}
To conduct a comprehensive comparison, we select the following state-of-the-art baselines for both ZS-CIR and CIR. We conducted the experiments by running these baselines on 8 × NVIDIA A800 GPUs.

\textbf{Baselines for ZS-CIR.} \textit{Captioning}: A method that employs the pre-trained captioning model BLIP-2 to generate captions of reference images and extract text features of the captions by CLIP text encoder to retrieve. \textit{Text-only}: Only the relative caption features, extracted by the CLIP text encoder, are utilized as retrieval features to calculate similarity for retrieval. \textit{Image-only}: Only the features of reference images, which are extracted by the CLIP image encoder, are used to compute similarity for retrieval. \textit{CIReVL}~\cite{karthik2024visionbylanguage}: A training-free method using a pre-trained generative VLM and asking an LLM to recompose the caption based on the textual target modification for retrieval. \textit{LDRE}~\cite{yang2024ldre}: A training-free method utilizing out-of-shelf tools to accurately retrieve a composed image based on a reference image and a relative caption. \textit{SEIZE}~\cite{yang2024semantic}: A novel method based on LDRE without training. \textit{PALAVRA}~\cite{eccv2022_palavra_cohen}: A two-stage training-dependent approach based textual inversion with a pre-trained mapping function and a subsequent optimization of the pseudo-word token. \textit{SEARLE}~\cite{baldrati2023zero}: A training-dependent method where pseudo-word tokens of unlabeled images are generated with an optimization-based textual inversion and then distill their knowledge to a textual inversion network. \textit{SEARLE-OTI}~\cite{agnolucci2024isearle}: A variant of SEARLE without the distillation network. \textit{Pic2Word}~\cite{saito2023pic2word}: A training-dependent method employs a pre-trained textual inversion network optimized by contrastive loss to capture the pseudo-word token, which is combined with the relative caption for retrieval. \textit{LinCIR}~\cite{gu2024lincir}: A language-only training framework for ZS-CIR.

\textbf{Baselines for CIR.} \textit{ARTEMIS}~\cite{delmas2022artemis}: A method based on two training jointly modules and their respective text-guided light-weight attention layers, where each handles one modality of the query. \textit{AMC}~\cite{zhu2023amc}: A method that leverages reinforcement learning to provide the model compression policy. \textit{VAL}~\cite{chen2020image}: A method fusing vision and language features via attention learning at varying representation depths. \textit{TIRG}~\cite{vo2019composing}: A method having the text modify the image feature via a gated residual connection. \textit{MAAF}~\cite{dodds2020modality}: A method using attention over undifferentiated text and image queries. \textit{DCNet}~\cite{kim2021dual}: A method that embeds the relationships between cells and their marker genes in the neural network, and can infer the cell landscape with more than 400 cell types based on bulk RNA-seq data. \textit{DWC}~\cite{huang2023dynamic}: A method to solve the inherent modality importance disparity, biased labeling noises, and modality gaps. \textit{CLIP4CIR}~\cite{baldrati2023composed}: A two-stage method that combines task-oriented fine-tuning with the training of a network, which can perform a fine-grained merging of the multimodal features.

\subsection{Evaluation Metrics}
\textbf{Mean Average Precision (mAP).} On ZeroSight, given that each query has multiple target images serving as ground truth,
we utilize mAP, a more fine-grained metric to take into account the rank of retrieval results. Firstly, for each positive target image, we calculate $P_i^j\text{@}j (1 \leq i \leq a, 1 \leq j \leq k)$
based on their rank positions as follows:
\begin{equation}
    P_i^j\text{@}j = \frac{i}{j},
\end{equation}
which represents the precision when the $i$-th positive target image is at the $j$-th rank position. Here, $a$ is the total number of positive target images.
Then, we apply Average Precision at k (AP@k) for each query, where $k$ represents the number of top-ranked retrieval results, as shown below:
\begin{equation}
   AP\text{@}k = \frac{\sum_{i=1}^{b} P_{i}^{j}\text{@}j}{\min(a,k)},
\end{equation}
where $b\ (0 \leq b \leq k)$ is the number of positive target images among the top $k$ retrieval results for each query.
Lastly, we calculate the mean AP@k as follows:
\begin{equation}\label{formula:map}
    mAP\text{@}k = \frac{\sum_{q=1}^{n} AP_{q}\text{@}k}{n},
\end{equation}
where  $AP_q\text{@}k$ represents the $AP\text{@}k$ of the retrieval results for the $q$-th query, and $n$ is the total number of queries.

\textbf{Positive-Negative Ranking mAP (PNR-mAP).}  
In the ZS-CIR task, negative target images refer to images that closely resemble positive target images but do not meet the relative caption to be classified as positive. 
Traditional evaluation metrics like mAP only consider the distribution of positive target images in the retrieval results, ignoring the presence and ranking of negative target images. The ranking order of negative target images relative to positive ones can also indicate retrieval quality. 
For example, two retrieval results with the same mAP value can differ significantly if, in one, negative target images are ranked before positive ones, while in the other, the opposite is true. The latter is clearly better, but mAP cannot distinguish between them. 
To address this, we propose the positive-negative ranking mAP (PNR-mAP) metric. Similar to mAP, we first calculate $P_i^j\text{@}j$ for each positive target image. Then, for the $i$-th positive target image at the $j$-th position, we compute the positive-negative ranking weight $w_i$ as follows:
\begin{equation}
w_i = 
\begin{cases} 
\frac{\sum_{x=1}^{l} \frac{N_x}{j}}{l} & \text{if } l \geq 1, \\
1 & \text{if } l = 0. 
\end{cases},
\end{equation}
where $l (0 \leq l < j)$ is the number of negative target images preceding the $i$-th positive target image at the $j$-th position, and $N_x (1 \leq N_x < j, 1 \leq x \leq l)$ denotes the rank position of the $x$-th negative target image. If more negative target images are ranked further ahead of a positive target image, its positive-negative ranking weight will be smaller. We then calculate the positive-negative ranking average precision at k by incorporating these weights, as follows:
\begin{equation}
   PNR\text{--}AP\text{@}k = \frac{\sum_{i=1}^{b} {w_i} \times {P_{i}^{j}\text{@}j}}{\min(a,k)},
\end{equation}
Lastly, with PNR-AP@k, we obtain the Positive-Negative Ranking Mean Average Precision at k, similar to the standard mAP@k calculation method, as shown in Eq. \ref{formula:map}.

\subsection{Results on Zerosight}
\textbf{Performance of ZS-CIR Methods on ZeroSight.} The evaluation results on ZeroSight, outlined in Table \ref{tab:ZS res}, provide a comprehensible comparison of ZS-CIR methods. Based on these results, we observed that:
(1) For the mAP metric, with different CLIP backbones, image-based methods achieve the best results, while methods based on image-to-text, such as LDRE, perform worse. This indicates that ZeroSight places more emphasis on the information contained within the image itself. 
(2) For the PNR-mAP metric, the performance for all methods are worser than their corresponding mAP performance.
This demonstrates that relying solely on image similarity is insufficient to correctly distinguish negative target images that closely resemble positive target images.
(3) In addition, training-based ZS-CIR methods outperform training-free methods on all backbones, which shows that learning based on image features during training can significantly improve the performance on ZeroSight.
(4) mAP and PNR-mAP are consistent in comparing method performance, but the average mAP of 19.94 across all methods is 22.93\% higher than the PNR-mAP of 16.22. This indicates that mAP does not account for the ranking of negative target images, leading to inflated retrieval performance and exaggerating the capabilities of CIR methods. This also demonstrates that our benchmark presents a challenging task.

\textbf{Performance of CIR Methods on ZeroSight.} We evaluate the performance of different CIR methods on our test set after training on our training set, as shown in Table \ref{tab:ses}. Among them, CLIP4CIR achieves the best performance under the PNR-mAP metric, while APTEMIS performs the worst under the same metric. 
Additionally, we compare the performance of these CIR methods on the FashionIQ validation set before and after training on our training set. Details are provided in the Sec.~\ref{more_ex:zs}.

\begin{table}[!t]
\caption{Performance of CIR Methods on the ZeroSight test set.}
\centering
\begin{tabular}{@{}cccccc@{}}
\toprule
\multirow{2}{*}{\textbf{Method}} & \multicolumn{5}{c}{\textbf{PNR-mAP@k}}                                       \\ \cmidrule(l){2-6} 
                                 & \textbf{k=5} & \textbf{k=10} & \textbf{k=25} & \textbf{k=50} & \textbf{Avg.} \\ \midrule
ARTEMIS\fontsize{6pt}{\baselineskip}\selectfont{(ICLR’2022)}               & 14.34        & 15.51         & 17.06         & 17.92         & 16.21         \\
AMC\fontsize{6pt}{\baselineskip}\selectfont{(TOMM’2023)}                   & 14.91        & 16.44         & 18.47         & 19.47         & 17.32         \\
VAL\fontsize{6pt}{\baselineskip}\selectfont{(CVPR’2020)}                   & 19.25        & 22.04         & 25.75         & 26.92         & 23.49         \\
TIRG\fontsize{6pt}{\baselineskip}\selectfont{(CVPR’2019)}                  & 20.43        & 23.52         & 26.63         & 27.17         & 24.44         \\
MAAF\fontsize{6pt}{\baselineskip}\selectfont{(ArXiv’2020)}                 & 20.16        & 24.72         & 28.33         & 29.84         & 25.76         \\
DCNet\fontsize{6pt}{\baselineskip}\selectfont{(AAAI’2021)}                 & 25.31        & 26.01         & 26.90         & 27.92         & 26.54         \\
DWC\fontsize{6pt}{\baselineskip}\selectfont{(AAAI’2024)}                   & 23.14        & 25.33         & 28.24         & 29.85         & 26.64         \\
CLIP4CIR\fontsize{6pt}{\baselineskip}\selectfont{(TOMM’2023)}              & \textbf{23.67}        & \textbf{29.23}         & \textbf{33.44}         & \textbf{34.93}         & \textbf{30.32}         \\ \bottomrule
\end{tabular}
\label{tab:ses}
\end{table}

\textbf{Performance of SC4CIR Method on ZeroSight.} To validate the effectiveness of SC4CIR, we conduct experiments using SEIZE for training-free methods and LinCIR for training-based methods. The experimental results are highlighted in Table \ref{tab:ZS res}. Based on these results, we can make the following observations:
(1) For mAP and PNR-mAP metrics, the average mAP of 25.47 across all methods represents a 5.90\% improvement over the previous average mAP of 24.05 and the average PNR-mAP of 22.11 across all methods is 12.86\% higher than the previous average PNR-mAP of 19.59. All methods show improvement when combined with SC4CIR, indicating its effectiveness. 
%
(2) For the mAP metric, the average result of 37.54 in all training-based methods is 1.79\% higher than the previous average result of 36.88 and the average result of 17.42 in training-free methods shows an improvement of 12.39\% over the previous average results of 15.50. For the PNR-mAP metric, the average result of 34.49 across all training-based methods shows an improvement of 6.35\% over the previous average result of 32.43 and the average result of 13.86 in training-free methods is 25.66\% higher than the previous average results of 11.03. Compared to training-based methods, training-free methods show greater improvement when using SC4CIR. This indicates that SC4CIR can help training-free methods gain the capabilities through training.
Results on common datasets are provided in the Sec.~\ref{model_copmare_common_datasets}.

\subsection{Results on Multi-target Datasets}
\label{model_copmare}

Since both CIRCO and ZeroSight are designed for one-to-many retrieval tasks and employ the same evaluation metric (mAP), we have also conducted experiments to compare the performance of different models on these two datasets, as shown in Table \ref{tab:circo&zerosight}. The results of the I2T method on ZeroSight are generally lower than those on CIRCO. Specifically, the mAP@5 result of \textit{SEIZE} on ZeroSight is $16.63$, which is a significant decrease of $48.77\%$ compared to the result of $32.46$ on CIRCO. On the other hand, the results of other methods on ZeroSight are generally higher than those on CIRCO. For instance, the mAP@5 result of \textit{LinCIR} on ZeroSight is $28.13$, which is a substantial increase of $55.24\%$ compared to the result of $12.59$ on CIRCO. This also reflects that CIRCO fails to balance the information from the reference image and the relative caption, allowing the I2T methods to achieve good results even after losing some reference image information, which is logically unacceptable. Furthermore, the results of other CIR methods without this loss process are improved on ZeroSight compared to CIRCO, which also indicates that ZeroSight better balances the information from the reference image and the relative caption (reference-target image consistency), providing a fairer reflection of the effectiveness of different methods.

\begin{table*}[t]
\renewcommand\arraystretch{0.8}
\centering
\small
\caption{Results of comparison among different models on CIRCO and ZeroSight test sets.}
\vspace{1mm}
\setlength{\tabcolsep}{5 pt}
\begin{tabular}{ccccccccccccc}
\toprule
\multirow{3}{*}{\textbf{Backbone}} & \multirow{3}{*}{\textbf{Method}} & \multirow{3}{*}{\textbf{Training-free}}  & \multicolumn{5}{c|}{\textbf{ZeroSight}}                                                           & \multicolumn{5}{c}{\textbf{CIRCO}}                                                                  \\ \cmidrule(l){4-13} 
                                   &                                  &                                         & \multicolumn{5}{c|}{\textbf{mAP@k}}                                                               & \multicolumn{5}{c}{\textbf{mAP@k}}                                       \\
                                   &                                  &                                         & \textbf{k=5} & \textbf{k=10} & \textbf{k=25} & \textbf{k=50} & \multicolumn{1}{c|}{\textbf{Avg.}} & \textbf{k=5} & \textbf{k=10} & \textbf{k=25} & \textbf{k=50} & \textbf{Avg.} \\ \midrule
\multirow{6}{*}{ViT-B/32}          & CIReVL\fontsize{6pt}{\baselineskip}\selectfont{(ICLR'2024)}                & \color{ForestGreen}{\CheckmarkBold}     & 7.76         & 8.51          & 9.57          & 10.12         & \multicolumn{1}{c|}{8.99}          & {14.94}    & {15.42}    & {17.00}    & {17.82}          & 16.30          \\
                                   & SEIZE\fontsize{6pt}{\baselineskip}\selectfont{(MM’2024)}                   & \color{ForestGreen}{\CheckmarkBold}     & 10.64        & 11.61         & 12.94         & 13.65         & \multicolumn{1}{c|}{12.21}         & {19.04} & {19.64} & {21.55} & {22.49}         & 20.68          \\
                                   & PALAVRA\fontsize{6pt}{\baselineskip}\selectfont{(ECCV'2022)}               & \color{red}{\XSolidBrush}               & 12.67        & 13.89         & 15.51         & 16.36         & \multicolumn{1}{c|}{14.61}         & 4.61           & 5.32           & 6.33           & 6.80         & 5.77         \\
                                   & SEARLE-OTI\fontsize{6pt}{\baselineskip}\selectfont{(ICCV'2023)}            & \color{red}{\XSolidBrush}               & 21.39        & 24.17         & 26.45         & 27.27         & \multicolumn{1}{c|}{24.82}         & 7.14           & 7.83           & 8.99           & 9.60         & 8.39         \\
                                   & SEARLE\fontsize{6pt}{\baselineskip}\selectfont{(ICCV'2023)}                & \color{red}{\XSolidBrush}               & 22.65        & 25.58         & 27.98         & 28.84         & \multicolumn{1}{c|}{26.26}         & 9.35           & 9.94           & 11.13          & 11.84         & 10.57         \\
                                   \rowcolor{gray!20} & SEIZE+SC4CIR\fontsize{6pt}{\baselineskip}\selectfont{(Ours)}                & \color{ForestGreen}{\CheckmarkBold}               & 12.59        & 13.71         & 15.21         & 16.12         & \multicolumn{1}{c|}{14.41}         & {22.16} & {22.89} & {24.08} & {24.59}         & 23.43         \\ \midrule
\multirow{10}{*}{ViT-L/14}         & Captioning                       & \color{ForestGreen}{\CheckmarkBold}     & 5.86         & 6.68          & 7.74          & 8.32          & \multicolumn{1}{c|}{7.15}          & 1.65           & 1.96           & 2.42           & 2.71          & 2.19         \\
                                   & Text-only                        & \color{ForestGreen}{\CheckmarkBold}     & 8.38         & 9.19          & 10.27         & 10.84         & \multicolumn{1}{c|}{9.67}          & 2.63           & 2.85           & 3.30           & 3.58          & 3.09          \\
                                   & CIReVL\fontsize{6pt}{\baselineskip}\selectfont{(ICLR'2024)}                & \color{ForestGreen}{\CheckmarkBold}     & 9.45         & 10.40         & 11.74         & 12.42         & \multicolumn{1}{c|}{11.00}         & {18.57}    & {19.01}    & {20.89}    & {21.80}          & 20.07          \\
                                   & Image-only                       & \color{ForestGreen}{\CheckmarkBold}     & 15.62        & 17.26         & 19.38         & 20.54         & \multicolumn{1}{c|}{18.20}         & 1.28           & 1.70           & 2.35           & 2.69         & 2.01          \\
                                   & SEIZE\fontsize{6pt}{\baselineskip}\selectfont{(MM’2024)}                   & \color{ForestGreen}{\CheckmarkBold}     & 13.02        & 14.36         & 15.98         & 16.80         & \multicolumn{1}{c|}{15.04}         & {24.98} & {25.82} & {28.24} & {29.35}         & 27.10         \\
                                   & Pic2Word\fontsize{6pt}{\baselineskip}\selectfont{(CVPR'2023)}              & \color{red}{\XSolidBrush}               & 15.06        & 18.09         & 19.78         & 20.21         & \multicolumn{1}{c|}{18.28}         & 8.72           & 9.51           & 10.64          & 11.29         & 10.04         \\
                                   & SEARLE-OTI\fontsize{6pt}{\baselineskip}\selectfont{(ICCV'2023)}            & \color{red}{\XSolidBrush}               & 24.93        & 30.68         & 33.89         & 34.71         & \multicolumn{1}{c|}{31.05}         & 10.18          & 11.03          & 12.72          & 13.67         & 11.90         \\
                                   & SEARLE\fontsize{6pt}{\baselineskip}\selectfont{(ICCV'2023)}                & \color{red}{\XSolidBrush}               & 26.12        & 32.17         & 35.55         & 36.41         & \multicolumn{1}{c|}{32.56}         & 11.68          & 12.73          & 14.33          & 15.12         & 13.47         \\
                                   & LinCIR\fontsize{6pt}{\baselineskip}\selectfont{(CVPR'2024)}                & \color{red}{\XSolidBrush}               & 28.13        & 34.48         & 38.03         & 38.93         & \multicolumn{1}{c|}{34.89}         & 12.59          & 13.58          & 15.00          & 15.85         & 14.26         \\
                                   \rowcolor{gray!20} & SEIZE+SC4CIR\fontsize{6pt}{\baselineskip}\selectfont{(Ours)}                & \color{ForestGreen}{\CheckmarkBold}               & 15.13        & 16.56         & 18.03         & 18.91         & \multicolumn{1}{c|}{17.16}         & {26.87} & {27.93} & {31.28} & {32.03}         & 29.53         \\\midrule
\multirow{6}{*}{ViT-G/14}          & CIReVL\fontsize{6pt}{\baselineskip}\selectfont{(ICLR'2024)}                & \color{ForestGreen}{\CheckmarkBold}     & 12.00        & 13.24         & 14.74         & 15.49         & \multicolumn{1}{c|}{13.87}        & {26.77}    & {27.59}    & {29.96}    & {31.03}         & 28.84          \\
                                   & SEIZE\fontsize{6pt}{\baselineskip}\selectfont{(MM’2024)}                   & \color{ForestGreen}{\CheckmarkBold}     & 16.63        & 18.37         & 20.47         & 21.54         & \multicolumn{1}{c|}{19.25}         & {32.46} & {33.77} & {36.46} & {37.55}         & 35.06         \\
                                   & Pic2Word\fontsize{6pt}{\baselineskip}\selectfont{(CVPR'2023)}              & \color{red}{\XSolidBrush}               & 16.07        & 19.39         & 21.25         & 21.73         & \multicolumn{1}{c|}{19.61}         & 5.54           & 5.59           & 6.68           & 7.12         & 6.23         \\
                                   & SEARLE\fontsize{6pt}{\baselineskip}\selectfont{(ICCV'2023)}                & \color{red}{\XSolidBrush}               & 31.46        & 37.95         & 41.72         & 41.83         & \multicolumn{1}{c|}{38.23}         & 13.20          & 13.85          & 15.32          & 16.04         & 14.60         \\
                                   & LinCIR\fontsize{6pt}{\baselineskip}\selectfont{(CVPR'2024)}                & \color{red}{\XSolidBrush}               & 32.00        & 38.57         & 42.38         & 42.51         & \multicolumn{1}{c|}{38.87}         & 19.71          & 21.01          & 23.13          & 24.18         & 22.01         \\
                                   \rowcolor{gray!20} & SEIZE+SC4CIR\fontsize{6pt}{\baselineskip}\selectfont{(Ours)}                & \color{ForestGreen}{\CheckmarkBold}               & 18.11        & 19.76         & 21.93         & 22.98         & \multicolumn{1}{c|}{20.70}         & {34.48} & {36.81} & {38.93} & {39.88}         & 37.53         \\\bottomrule
\end{tabular}
\label{tab:circo&zerosight}
\vspace{-3mm}
\end{table*}

\subsection{Performance of SC4CIR on Common Datasets}
\label{model_copmare_common_datasets}

\textbf{CIRCO:} 
The left section of Table \ref{tab:circo&cirr} displays CIRCO test results. Based on them, we have the following observations: 
(1) Among the simpler baselines, \textit{Image-only} and \textit{Captioning} perform worse than \textit{Text-only}, indicating the importance of relative text for CIR. Besides, \textit{Captioning} outperforms \textit{Image-only}, suggesting that textual features from image captions are more suitable for CIR than direct visual features.
(2) Among I2T-based methods, the methods based on pre-trained captioning models, \textit{CIReVL} and \textit{SEIZE}, perform better than the methods based on pseudo-word, \textit{PALAVRA}, \textit{Pic2Word}, \textit{SEARLE}, and \textit{LinCIR}. This demonstrates that the captions generated by the captioning model possess semantics that are more suitable for CLIP text encoding compared to textual inversion.
(3) In the absence of SC4CIR, \textit{SEIZE} already outperforms all baselines across all metrics and CLIP backbones. When SC4CIR is applied, the performance improves even further. For instance, using ViT-L/14, \textit{SEIZE with SC4CIR} outperforms \textit{SEIZE} by 7.57\% in mAP@5, 8.17\% in mAP@10, 10.76\% in mAP@25, and 9.13\% in mAP@50, proving the effectiveness of SC4CIR for CIR methods.

\textbf{CIRR:}
The right section of Table \ref{tab:circo&cirr} presents CIRR test results.
From these results, key observations include: 
(1) 
\textit{Text-only} performs significantly better than \textit{Image-only} and \textit{Captioning}, indicating a minimal correlation between reference and target images due to the noisy dataset and the lesser information provided by reference images.
(2) Despite the noise, SC4CIR can continuously improve the performance of \textit{SEIZE} and \textit{SEIZE with SC4CIR} outperforms all baselines across all CLIP backbones.
This highlights the robustness and adaptability of SC4CIR even in noisy data and diverse scenarios.
(3) Using ViT-L/14 CLIP, \textit{SEIZE with SC4CIR} surpasses the second-best method \textit{SEIZE} by 7.82\% in Recall@1, 5.56\% in Recall@5, and 3.80\% in Recall@10, emphasizing the effectiveness of SC4CIR.

\textbf{FashionIQ:}
As shown in Table \ref{tab:new_fashioniq}, we can have the following observations: (1) \textit{SC4CIR} can consistently improve the performance of \textit{LinCIR} across all three subsets, demonstrating the effectiveness of \textit{SC4CIR} on Fashion-IQ. (2) \textit{LinCIR+SC4CIR} performs best on the R@10 metric of the Shirt subset and all metrics of the Toptee subset, even surpassing the performance of the trained CIR method \textit{CASE} on these metrics. Additionally, \textit{LinCIR+SC4CIR}'s performance on the R@50 metric of the Shirt subset and all metrics of the Dress subset is second only to \textit{CASE}, proving that \textit{SC4CIR} can consistently enhance the effectiveness of CIR methods and even help ZS-CIR methods achieve performance comparable to trained CIR methods to some extent.

\begin{table*}[t]
\renewcommand\arraystretch{0.8}
\centering
\small
\caption{Results of comparison among different models on CIRCO and CIRR test sets. \rm Best scores are highlighted in bold. 
}
\vspace{1mm}
\setlength{\tabcolsep}{5pt} 
\begin{tabular}{clc|cccc|cccccc}
\toprule
\multirow{3}{*}{\textbf{Backbone}} & \multicolumn{1}{c}{\multirow{3}{*}{\textbf{Method}}}          & \multirow{3}{*}{\textbf{Training-free}} & \multicolumn{4}{c|}{\textbf{CIRCO}}                               & \multicolumn{6}{c}{\textbf{CIRR}}                                                                                        \\ \cmidrule(l){4-13} 
                                   & \multicolumn{1}{c}{}                                          &                                         & \multicolumn{4}{c|}{\textbf{mAP@k}}                               & \multicolumn{3}{c|}{\textbf{Recall@k}}                                & \multicolumn{3}{c}{\textbf{Rs@k}}                \\
                                   & \multicolumn{1}{c}{}                                          &                                         & \textbf{k=5}   & \textbf{k=10}  & \textbf{k=25}  & \textbf{k=50}  & \textbf{k=1}   & \textbf{k=5}   & \multicolumn{1}{c|}{\textbf{k=10}}  & \textbf{k=1}   & \textbf{k=2}   & \textbf{k=3}   \\ \midrule
\multirow{5}{*}{ViT-B/32}          & PALAVRA\fontsize{6pt}{\baselineskip}\selectfont{(ECCV'22)}    & \color{red}{\XSolidBrush}               & 4.61           & 5.32           & 6.33           & 6.80           & 16.62          & 43.49          & \multicolumn{1}{c|}{58.51}          & 41.61          & 65.30          & 80.94          \\
                                   & SEARLE\fontsize{6pt}{\baselineskip}\selectfont{(ICCV'23)}     & \color{red}{\XSolidBrush}               & 9.35           & 9.94           & 11.13          & 11.84          & 24.00          & {53.42}    & \multicolumn{1}{c|}{{66.82}}    & 54.89          & 76.60          & 88.19          \\
                                   & SEARLE-OTI\fontsize{6pt}{\baselineskip}\selectfont{(ICCV'23)} & \color{red}{\XSolidBrush}               & 7.14           & 7.83           & 8.99           & 9.60           & {24.27}    & 53.25          & \multicolumn{1}{c|}{66.10}          & 54.10          & 75.81          & 87.33          \\
                                   & CIReVL\fontsize{6pt}{\baselineskip}\selectfont{(ICLR'24)}     & \color{ForestGreen}{\CheckmarkBold}     & {14.94}    & {15.42}    & {17.00}    & {17.82}    & 23.94          & 52.51          & \multicolumn{1}{c|}{66.00}          & {60.17}    & {80.05}    & {90.19}    \\
                                   & SEIZE\fontsize{6pt}{\baselineskip}\selectfont{(MM'24)}          & \color{ForestGreen}{\CheckmarkBold}     & \textbf{19.04} & \textbf{19.64} & \textbf{21.55} & \textbf{22.49} & \textbf{27.47} & \textbf{57.42} & \multicolumn{1}{c|}{\textbf{70.17}} & \textbf{65.59} & \textbf{84.48} & \textbf{92.77} \\
                                   \rowcolor{gray!20}& SEIZE+SC4CIR\fontsize{6pt}{\baselineskip}\selectfont{(Ours)}          & \color{ForestGreen}{\CheckmarkBold}     & \textbf{22.16} & \textbf{22.89} & \textbf{24.08} & \textbf{24.59} & \textbf{29.58} & \textbf{59.39} & \multicolumn{1}{c|}{\textbf{73.16}} & \textbf{68.36} & \textbf{85.61} & \textbf{93.48} \\
                                   \midrule
\multirow{9}{*}{ViT-L/14}          & Image-only                                                    & \color{ForestGreen}{\CheckmarkBold}     & 1.28           & 1.70           & 2.35           & 2.69           & 3.64           & 12.75          & \multicolumn{1}{c|}{23.32}          & 11.58          & 31.41          & 45.26          \\
                                   & Text-only                                                     & \color{ForestGreen}{\CheckmarkBold}     & 2.63           & 2.85           & 3.30           & 3.58           & 20.51          & 43.21          & \multicolumn{1}{c|}{55.08}          & {60.39}    & {80.02}    & {90.05}    \\
                                   & Captioning                                                    & \color{ForestGreen}{\CheckmarkBold}     & 1.65           & 1.96           & 2.42           & 2.71           & 4.05           & 15.88          & \multicolumn{1}{c|}{25.69}          & 20.87          & 40.60          & 60.89          \\
                                   & Pic2Word\fontsize{6pt}{\baselineskip}\selectfont{(CVPR'23)}   & \color{red}{\XSolidBrush}               & 8.72           & 9.51           & 10.64          & 11.29          & 23.90          & 51.70          & \multicolumn{1}{c|}{65.30}          & 53.76          & 74.46          & 87.08          \\
                                   & SEARLE\fontsize{6pt}{\baselineskip}\selectfont{(ICCV'23)}     & \color{red}{\XSolidBrush}               & 11.68          & 12.73          & 14.33          & 15.12          & 24.24          & 52.48          & \multicolumn{1}{c|}{66.29}          & 53.76          & 75.01          & 88.19          \\
                                   & SEARLE-OTI\fontsize{6pt}{\baselineskip}\selectfont{(ICCV'23)} & \color{red}{\XSolidBrush}               & 10.18          & 11.03          & 12.72          & 13.67          & 24.87          & 52.31          & \multicolumn{1}{c|}{66.29}          & 53.80          & 74.31          & 86.94          \\
                                   & LinCIR\fontsize{6pt}{\baselineskip}\selectfont{(CVPR'24)}     & \color{red}{\XSolidBrush}               & 12.59          & 13.58          & 15.00          & 15.85          & {25.04}    & {53.25}    & \multicolumn{1}{c|}{{66.68}}    & 57.11          & 77.37          & 88.89          \\
                                   & CIReVL\fontsize{6pt}{\baselineskip}\selectfont{(ICLR'24)}     & \color{ForestGreen}{\CheckmarkBold}     & {18.57}    & {19.01}    & {20.89}    & {21.80}    & 24.55          & 52.31          & \multicolumn{1}{c|}{64.92}          & 59.54          & 79.88          & 89.69          \\
                                   & SEIZE\fontsize{6pt}{\baselineskip}\selectfont{(MM'24)}          & \color{ForestGreen}{\CheckmarkBold}     & \textbf{24.98} & \textbf{25.82} & \textbf{28.24} & \textbf{29.35} & \textbf{28.65} & \textbf{57.16} & \multicolumn{1}{c|}{\textbf{69.23}} & \textbf{66.22} & \textbf{84.05} & \textbf{92.34} \\
                                   \rowcolor{gray!20}& SEIZE+SC4CIR\fontsize{6pt}{\baselineskip}\selectfont{(Ours)}          & \color{ForestGreen}{\CheckmarkBold}     & \textbf{26.87} & \textbf{27.93} & \textbf{31.28} & \textbf{32.03} & \textbf{30.89} & \textbf{60.34} & \multicolumn{1}{c|}{\textbf{71.86}} & \textbf{68.43} & \textbf{87.31} & \textbf{94.99} \\ \midrule
\multirow{5}{*}{ViT-G/14}          & Pic2Word\fontsize{6pt}{\baselineskip}\selectfont{(CVPR'23)}   & \color{red}{\XSolidBrush}               & 5.54           & 5.59           & 6.68           & 7.12           & 30.41          & 58.12          & \multicolumn{1}{c|}{69.23}          & {68.92}    & {85.45}    & 93.04          \\
                                   & SEARLE\fontsize{6pt}{\baselineskip}\selectfont{(ICCV'23)}     & \color{red}{\XSolidBrush}               & 13.20          & 13.85          & 15.32          & 16.04          & 34.80          & 64.07          & \multicolumn{1}{c|}{75.11}          & 68.72          & 84.70          & {93.23}    \\
                                   & LinCIR\fontsize{6pt}{\baselineskip}\selectfont{(CVPR'24)}     & \color{red}{\XSolidBrush}               & 19.71          & 21.01          & 23.13          & 24.18          & {35.25}    & {64.72}    & \multicolumn{1}{c|}{{76.05}}    & 63.35          & 82.22          & 91.98          \\
                                   & CIReVL\fontsize{6pt}{\baselineskip}\selectfont{(ICLR'24)}     & \color{ForestGreen}{\CheckmarkBold}     & {26.77}    & {27.59}    & {29.96}    & {31.03}    & 34.65          & 64.29          & \multicolumn{1}{c|}{75.06}          & 67.95          & 84.87          & 93.21          \\
                                   & SEIZE\fontsize{6pt}{\baselineskip}\selectfont{(MM'24)}          & \color{ForestGreen}{\CheckmarkBold}     & \textbf{32.46} & \textbf{33.77} & \textbf{36.46} & \textbf{37.55} & \textbf{38.87} & \textbf{69.42} & \multicolumn{1}{c|}{\textbf{79.42}} & \textbf{74.15} & \textbf{89.23} & \textbf{95.71} \\
                                   \rowcolor{gray!20}& SEIZE+SC4CIR\fontsize{6pt}{\baselineskip}\selectfont{(Ours)}          & \color{ForestGreen}{\CheckmarkBold}     & \textbf{34.48} & \textbf{36.81} & \textbf{38.93} & \textbf{39.88} & \textbf{40.23} & \textbf{71.58} & \multicolumn{1}{c|}{\textbf{81.68}} & \textbf{76.63} & \textbf{90.35} & \textbf{96.88} \\
                                   \bottomrule
\end{tabular}

\label{tab:circo&cirr}
\end{table*}

\begin{table*}[t]
\centering
\caption{Results of comparison among different models on FashionIQ.}
\label{tab:results}
\small 
\setlength{\tabcolsep}{10 pt}
\begin{tabular}{@{}ccccccccc@{}}
\toprule
Methods & \multicolumn{2}{c}{Shirt} & \multicolumn{2}{c}{Dress} & \multicolumn{2}{c}{Toptee} & \multicolumn{2}{c}{Average} \\
\cmidrule(lr){2-3} \cmidrule(lr){4-5} \cmidrule(lr){6-7} \cmidrule(lr){8-9}
 & R@10 & R@50 & R@10 & R@50 & R@10 & R@50 & R@10 & R@50 \\
\midrule
CASE & 48.48 & 70.23 & 47.44 & 69.36 & 50.18 & 72.24 & 48.79 & 70.68 \\
TransAgg(Laion-CIR-Combined) & - & - & - & - & - & 34.36 & 55.13 & - \\
CoVR & - & - & - & - & - & 27.70 & 44.63 & - \\
LinCIR(ViT-G/14) & 46.76 & 65.11 & 38.08 & 60.88 & 50.48 & 71.09 & 45.11 & 65.69 \\
LinCIR(ViT-G/14)+SC4CIR & 49.34 & 68.02 & 41.53 & 63.91 & 53.18 & 74.33 & 48.02 & 68.75 \\
\bottomrule
\end{tabular}
\label{tab:new_fashioniq}
\end{table*}

\subsection{Performance of ZS-CIR Methods after Fine-tuning on ZeroSight}
\label{more_ex:zs}
\textbf{FashionIQ} Table \ref{tab:fashioniq} shows the performance of these CIR methods on the FashionIQ validation set before and after fine-tuning on our training set. Based on the results, we have following observations: (1) After fine-tuning on our dataset, all methods exhibit retrieval capabilities that are either comparable to or improved from their pre-training performance. This indicates that our dataset, as a ZS-CIR dataset, can provide training data for CIR methods. (2) Among the methods, CLIP4CIR shows the highest improvement after fine-tuning, with a 3.13\% increase in average Recall@50. This demonstrates the effectiveness of the training data provided by ZeroSight.

\textbf{CIRR} The experimental results, outlined in Table \ref{tab:cirr}, show the performance of these CIR methods on the CIRR test set before and after fine-tuning on our training set. Based on the results, we can make the following observations: (1) After fine-tuning on our dataset, all methods also can exhibit retrieval capabilities that are either comparable to or improved from their pre-training performance. This indicates that our dataset, as a ZS-CIR dataset, can provide training data for CIR methods. (2) Among the methods, as the same as results on the FashionIQ validation set, CLIP4CIR shows the highest improvement after fine-tuning, with a 4.65\% increase in average Recall@50. This further demonstrates the effectiveness of the training data provided by our dataset.

\begin{table*}[t]
\renewcommand\arraystretch{0.8}
\centering
\small
\setlength{\tabcolsep}{9 pt}
\caption{\small Comparison results of different models before and after fine-tuning on ZeroSight using FashionIQ validation set. The best scores are highlighted in bold.
}
\vspace{1mm}
\begin{tabular}{@{}clclclclclclclclcl@{}}
\toprule
\multirow{2}{*}{Model}               & \multicolumn{1}{c}{\multirow{2}{*}{Stage}} & \multicolumn{2}{c}{Shirt}       & \multicolumn{2}{c}{Dress}       & \multicolumn{2}{c}{Toptee}      & \multicolumn{2}{c}{Average}     \\ \cmidrule(l){3-10} 
                                     & \multicolumn{1}{c}{}                       & R@10           & R@50           & R@10           & R@50           & R@10           & R@50           & R@10           & R@50           \\ \midrule
\multirow{2}{*}{MAAF(ArXiv’2020)}    & Pre-train                                  & 19.68          & \textbf{35.92} & 15.32          & 34.75          & 21.01          & 41.36          & 18.67          & 37.34          \\
                                     & Fine-tune                                  & \textbf{20.07} & 35.38          & \textbf{16.81} & \textbf{35.60} & \textbf{21.98} & \textbf{42.78} & \textbf{19.62} & \textbf{37.92} \\ \midrule
\multirow{2}{*}{CLIP4CIR(TOMM’2023)} & Pre-train                                  & 25.32          & 41.46          & 20.92          & 40.70          & 27.49          & 47.53          & 24.58          & 43.23          \\
                                     & Fine-tune                                  & \textbf{26.84} & \textbf{45.14} & \textbf{21.47} & \textbf{43.43} & \textbf{28.35} & \textbf{49.72} & \textbf{25.55} & \textbf{46.10} \\ \midrule
\multirow{2}{*}{ARTEMIS(ICLR’2022)}  & Pre-train                                  & 16.25          & \textbf{28.79} & 8.39           & 21.99          & 14.37          & \textbf{29.07} & 13.00          & \textbf{26.62} \\
                                     & Fine-tune                                  & \textbf{16.81} & 28.24          & \textbf{9.65}  & \textbf{22.05} & \textbf{15.10} & 28.50          & \textbf{13.85} & 26.26          \\ \midrule
\multirow{2}{*}{AMC(TOMM’2023)}      & Pre-train                                  & 15.66          & 26.97          & 8.57           & 20.84          & 13.96          & 27.23          & 12.73          & 25.01          \\
                                     & Fine-tune                                  & \textbf{16.19} & \textbf{27.16} & \textbf{9.31}  & \textbf{21.22} & \textbf{14.54} & \textbf{27.41} & \textbf{13.35} & \textbf{25.26} \\ \bottomrule
\end{tabular}

\label{tab:fashioniq}

\end{table*}

\begin{table*}[t]
\centering
\normalsize
\setlength{\tabcolsep}{13 pt}
\caption{\small Comparison results of different models before and after fine-tuning on ZeroSight using CIRR test set. The best scores are highlighted in bold.}
\vspace{1mm}
\begin{tabular}{@{}cccccccc@{}}
\toprule
\multirow{2}{*}{Model}               & \multirow{2}{*}{Stage} & \multicolumn{6}{c}{Recall@k}                                                                        \\
                                     &                        & k=1            & k=5            & k=10           & k=50           & k=100          & Avg.           \\ \midrule
\multirow{2}{*}{MAAF(ArXiv’2020)}    & Pre-train              & 20.99          & 31.04          & 49.21          & 62.27          & \textbf{87.42} & 50.18          \\
                                     & Fine-tune              & \textbf{22.53} & \textbf{32.87} & \textbf{49.95} & \textbf{63.28} & 86.58          & \textbf{51.04} \\ \midrule
\multirow{2}{*}{CLIP4CIR(TOMM’2023)} & Pre-train              & \textbf{6.17}  & 22.53          & 44.58          & 59.74          & 86.48          & 43.90          \\
                                     & Fine-tune              & 5.83           & \textbf{25.30} & \textbf{49.86} & \textbf{65.52} & \textbf{89.76} & \textbf{47.25} \\ \midrule
\multirow{2}{*}{ARTEMIS(ICLR’2022)}  & Pre-train              & 11.36          & 27.25          & 41.56          & 57.09          & 84.00          & 44.25          \\
                                     & Fine-tune              & \textbf{12.49} & \textbf{29.97} & \textbf{45.71} & \textbf{62.79} & \textbf{86.39} & \textbf{47.47} \\ \bottomrule
\end{tabular}

\label{tab:cirr}

\end{table*}

\subsection{Performance of Various Query Categories}
The evaluation results, as shown in Table \ref{tab:query}, demonstrate the performance of these CIR methods for six different categories of queries. Based on these results, we have the following observations: (1) For queries of the Attribute Change category, all methods perform relatively well, indicating that queries of the Attribute Change category are relatively simple for current CIR methods. (2) Conversely, for queries of the Background Change category, all methods have poor performances, suggesting that queries of the Background Change category in our dataset are challenging for current CIR methods. (3) Overall, all results of all methods across these six categories of queries are not very well, indicating that our dataset remains challenging for existing CIR methods.

\begin{table*}[t]
\renewcommand\arraystretch{0.9}
\centering
\small
\setlength{\tabcolsep}{16 pt}
\caption{Performance of Various Query Categories. \textbf{Add.}: Addition. \textbf{Sub.}: Subtraction. \textbf{View.}: Viewpoint Change. \textbf{Back.}: Background Change. \textbf{Att.}: Attribute Change. \textbf{Rel.}: Relative Statement. \textbf{Avg.}: Average.}
\vspace{1mm}

\begin{tabular}{@{}cccccccc@{}}
\toprule
\textbf{Method}     & \textbf{Add.} & \textbf{Sub.} & \textbf{View.} & \textbf{Back.} & \textbf{Att.} & \textbf{Rel.} & \textbf{Avg.} \\ \midrule
TIRG(CVPR’2019)     & 23.81             & 21.65                & 24.19                     & 22.01                      & 26.17                     & 22.97                       & 23.52         \\
VAL(CVPR’2020)      & 22.38             & 20.22                & 22.68                     & 20.62                      & 24.54                     & 21.51                       & 22.04         \\
DCNet(AAAI’2021)    & 26.23             & 24.05                & 26.73                     & 24.35                      & 28.91                     & 25.44                       & 26.01         \\
MAAF(ArXiv’2020)    & 24.98             & 22.81                & 25.41                     & 23.14                      & 27.49                     & 24.16                       & 24.72         \\
ARTEMIS(ICLR’2022)  & 16.04             & 13.94                & 16.02                     & 14.48                      & 17.36                     & 15.04                       & 15.51         \\
CLIP4CIR(TOMM’2023) & 29.35             & 27.15                & 30.01                     & 27.38                      & 32.45                     & 28.63                       & 29.23         \\ \bottomrule
\end{tabular}

\label{tab:query}

\end{table*}

\subsection{Ablation Study}
In this section, we conduct ablation studies to evaluate the individual contributions of components in our method. We select the SEIZE method with a ViT-L backbone as the baseline to evaluate the impact of our SC4CIR on its performance on the CIRCO test set. The results of the ablation study are shown in Table \ref{tab:abl}.


\textbf{Impact of Reverse Process 1.}
Compared to Reverse Process 2, it exhibits improvement but with a relatively limited margin. This is because Reverse Process 1, despite serving as a reverse process for verification, fundamentally relies on the original architecture of the method for CIR operations. It merely maximizes the utilization of inherent capabilities of the baseline method, thus its effectiveness remains constrained by the original architecture of the method and initial retrieval results.

\begin{table}[!t]
\caption{\small Ablation of SC4CIR on CIRCO test set.}
\centering
\begin{tabular}{@{}cccccc@{}}
\toprule
\multirow{2}{*}{\textbf{Method}} & \multicolumn{5}{c}{\textbf{mAP@k}}                                       \\ \cmidrule(l){2-6} 
                                 & \textbf{k=5} & \textbf{k=10} & \textbf{k=25} & \textbf{k=50} & \textbf{Avg.} \\ \midrule
SEIZE\fontsize{6pt}{7.2pt}               & 24.98        & 25.82         & 28.24         & 29.35         & 27.10         \\
SEIZE + Reverse Process 1\fontsize{6pt}{7.2pt}                  & 25.07        & 26.64         & 29.44         & 30.42         & 27.89         \\
SEIZE + Reverse Process 2\fontsize{6pt}{7.2pt}                   & 25.78        & 26.93         & 30.24         & 30.64         & 28.40         \\
SEIZE + SC4CIR\fontsize{6pt}{7.2pt}                  & \textbf{26.87}        & \textbf{27.93}         & \textbf{31.28}         & \textbf{32.03}         & \textbf{29.53}           \\ \bottomrule
\end{tabular}
  \label{tab:abl}
\end{table}

\textbf{Impact of Reverse Process 2.}
The results demonstrates significantly greater improvement over Reverse Process 1. This shows that: (1) Unlike Reverse Process 1, Reverse Process 2 bypasses the limitations of the original method by leveraging the full capacity of Large Vision-Language Models (LVLMs, e.g., GPT-4o), enabling superior performance unbound by the constraints of the baseline method. (2) The LVLM exploits complete visual information from both reference and target images to refine ranking, underscoring that the original method does not make the full use of visual cues in reference images. (3) This under-utilization of visual information further demonstrates the necessity of constructing a dataset that emphasizes the visual information contained within images.

\subsection{Analysis of SC4CIR Efficiency}
\label{SC4CIR_Efficiency}
Firstly, even when scaling up existing methods at the same time cost, the performance of these methods may not necessarily improve. This is because the parameter settings given by existing methods are typically the result of numerous experiments by their authors, aiming to achieve a good balance between performance and cost. 

Taking \textit{SEIZE} (Semantic Editing Increment Benefits Zero-Shot Composed Image Retrieval) as an example, the left curve in Figure 5 of the paper shows the impact of the number of multiple captions on the performance of \textit{SEIZE} on CIRCO, which can be converted into a table as Table \ref{tab:sc4cir_expla}, where N is the number of multiple captions. As shown in the table, when N increases from 10 to 14, the performance of \textit{SEIZE} shows a slight improvement. However, when N further increases to 20, the performance of \textit{SEIZE} does not improve significantly. Increasing N from 10 to 20 essentially doubles the size of one of modules of \textit{SEIZE}, but the effect is not as good as directly using \textit{SC4CIR} when $N=10$. The mAP@25 result of $31.28$ for SC4CIR is even $11.4\%$ higher compared to the result of $28.1$ for $N=20$. This reflects that simply expanding the size of \textit{SEIZE} does not necessarily improve its performance, and it will be better to combine it with \textit{SC4CIR}.

\par
Secondly, while \textit{SC4CIR} increases inference time, it provides stable and significant performance gains, evidenced by improvements of $29.39\%$ (\textit{SEIZE}) and $6.86\%$ (\textit{LinCIR}) in Avg PNR-mAP. Crucially, \textit{SC4CIR} is a plug-and-play module, distinct in design and function from existing methods. It should therefore be evaluated on its ability to enhance state-of-the-art techniques, not solely on raw efficiency. Furthermore, its parameter N offers flexibility: users can tune it to balance effectiveness and computational cost based on specific application requirements.

\begin{table}[!t]
\caption{\small Performance Comparison of SEIZE Varying Numbers of Captions and SEIZE Integrated with SC4CIR on the CIRCO Test Set.}
\centering
\footnotesize
\begin{tabular}{@{}cccccc@{}}
\toprule
\multirow{2}{*}{\textbf{Method}} & \multicolumn{5}{c}{\textbf{mAP@k}}                                       \\ \cmidrule(l){2-6} 
                                 & \textbf{k=5} & \textbf{k=10} & \textbf{k=25} & \textbf{k=50} & \textbf{Avg.} \\ \midrule
N=5\fontsize{6pt}{7.2pt}               & 24.27        & 25.04         & 27.54         & 28.59         & 26.36         \\
N=10\fontsize{6pt}{7.2pt}                  & 24.59        & 25.14         & 27.97         & 29.03         & 26.68         \\
N=14\fontsize{6pt}{7.2pt}                   & 25.11        & 25.95         & 28.10         & 29.12         & 27.07         \\
N=20\fontsize{6pt}{7.2pt}                   & 24.95        & 25.92         & 28.12         & 29.16         & 27.04         \\
N=10\&\&+SC4CIR\fontsize{6pt}{7.2pt}                  & {26.89}        & {27.93}         & {31.28}         & {32.03}         & {29.53}           \\ \bottomrule
\end{tabular}
\label{tab:sc4cir_expla}
\end{table}

\subsection{Case Study}
\label{sec:case_study}

\begin{figure}[!t]
\centering
\vspace{0.2cm}
\includegraphics[width=3.5in]{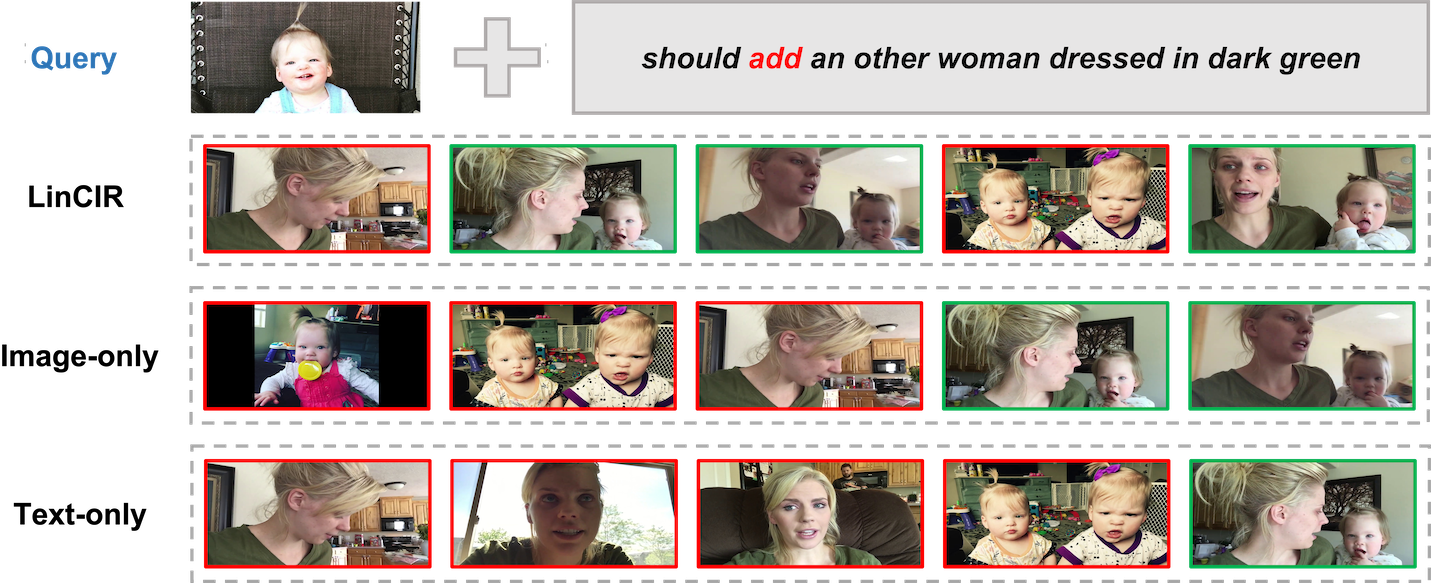}
\caption{\textbf{Case study of our ZeroSight dataset.}}
\label{fig:case_study}
    \vspace{-5mm}
\end{figure}

We select three representative methods, Text-only, Image-only and LinCIR, to compare their top-5 retrieval results on ZeroSight, with details in the Figure \ref{fig:case_study}. Images with red borders represent negative target images, while images with green borders represent positive target images. This comparison shows that ZeroSight is a challenge to the existing ZS-CIR methods.

\subsection{Visualization}   

\begin{figure*}[!t]
\centering
\includegraphics[width=0.95\textwidth]{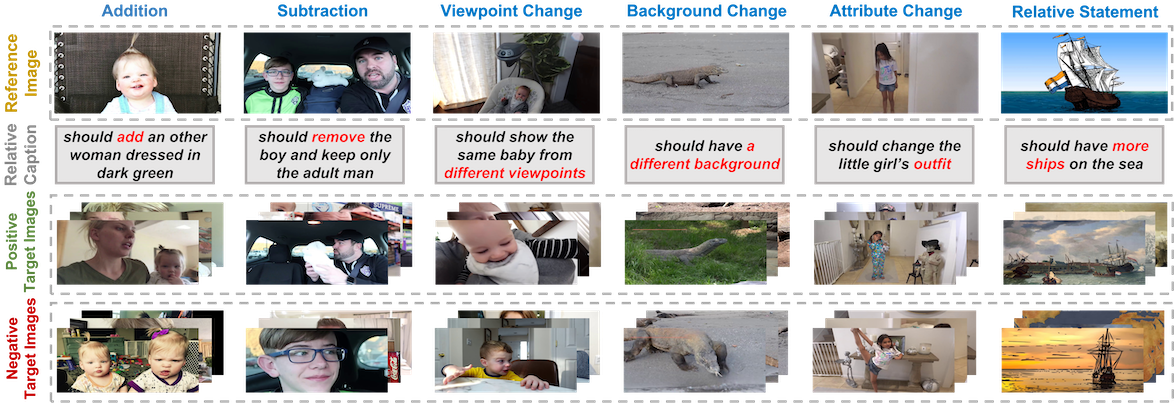}
\captionsetup{width=0.95\textwidth}
\caption{\small \textbf{Examples of our ZS-CIR dataset.} We divide all queries into six categories, including Addition, Subtraction, Viewpoint Change, Background Change, Attribute Change and Relative Statement. The words in relative caption, highlighted in red, indicate the core characteristic of the category to which the query belongs.}
\label{fig:examples visualization}
\end{figure*}

Figure \ref{fig:examples visualization} shows examples of our dataset, which contain queries with corresponding reference image, relative caption, positive target images and negative target images. In addition, to ensure a rich variety of queries in our dataset, we further divide all queries into six categories, including Addition, Subtraction, Viewpoint Change, Background Change, Attribute Change and Relative Statement. For example, if a relative caption contains the content ``should add another woman dressed in dark green,'' the corresponding query should be classified as category ``Addition.'' The reason is that the relative caption intends to add new content to its corresponding reference image and then perform retrieval based on this compose content. Furthermore, in this relative caption, the word ``add'' also reflects the core characteristic of category ``Addition.''

\subsection{Prompts In Construction Pipeline}
\label{prompt}
Tab~\ref{tab:prompt_1} and Tab~\ref{tab:prompt_2} show prompts used for selection of candidate reference images. Tab~\ref{tab:prompt_3} and Tab~\ref{tab:prompt_4} respectively show prompts used for generation of candidate relative captions and final relative captions within our dataset by both the LVLM (such as GPT-4o) and the LLM (such as GPT-4) mentioned. 

\begin{table}[!htb]
\caption{\small Prompt for Selecting Candidate Reference Images (When Selecting the First Candidate Reference Image).}
\begin{tcolorbox}[
        colback=gray!10,
        colframe=black,
        width=\columnwidth,  
        arc=2mm, 
        auto outer arc,
        title={Prompt for selecting candidate reference images (when selecting the first candidate reference image)}]		
        \footnotesize  
        
        Task Description:

        You will be given several images extracted from video frames in sequence. You need to select one image that meets the following requirements.

        Requirements:
        
        1. The selected image must be aesthetically pleasing, conforming to human aesthetics, and must be clear without any blurry areas.
        
        2. The selected image should be sufficiently distinct from the other images, and the difference should be describable in one sentence.
        
        3. The number of people and objects in the image should be moderate, neither too many nor too few.
        
        4. The overall arrangement of elements in the image should not appear too monotonous or too chaotic.
        
        5. Images with only English words are not allowed to be selected.
        
        6. Only use Arabic numerals to output the sequence number of the selected image, and do not output any other characters.
        
        7. If there is no suitable image, output the number 0 using Arabic numerals, and do not output any other characters.

        \{Input image sequence\}

\end{tcolorbox}
\label{tab:prompt_1}
\end{table}

\begin{table}[!htb]
\caption{\small Prompt for Selecting Candidate Reference Images (When Selecting Other Candidate Reference Images).}
\begin{tcolorbox}[
        colback=gray!10,
        colframe=black,
        width=\columnwidth,  
        arc=2mm, 
        auto outer arc,
        title={Prompt for selecting candidate reference images (when selecting other candidate reference images)}]		
        \footnotesize  
        
        Task Description:

        You will be given several images extracted from video frames in sequence. The first image will be used as a reference image, and you need to select one image from the remaining images that meets the following requirements.

        Requirements:
        
        1. The selected image must not resemble the reference image and must have sufficient differences from it.
        
        2. The selected image must be aesthetically pleasing, conforming to human aesthetics, and must be clear without any blurry areas.
        
        3. The selected image should be sufficiently distinct from the other images (excluding the reference image), and the difference should be describable in one sentence.
        
        4. The number of people and objects in the image should be moderate, neither too many nor too few.
        
        5. The overall arrangement of elements in the image should not appear too monotonous or too chaotic.
        
        6. Images with only English words are not allowed to be selected.
        
        7. Only use Arabic numerals to output the sequence number of the selected image, ranging from 2 to 10, and do not output any other characters.
        
        8. If there is no suitable image, output the number 1 using Arabic numerals, and do not output any other characters.

        \{Input image sequence\}

\end{tcolorbox}
\label{tab:prompt_2}
\end{table}

\begin{table}[!htb]
\caption{\small Prompt for Generating Candidate Relative Captions.}
\begin{tcolorbox}[
        colback=gray!10,
        colframe=black,
        width=\columnwidth,  
        arc=2mm, 
        auto outer arc,
        title={Prompt for generating candidate relative captions}]		
        \footnotesize  
        
        Task Description:

        You will be given two images in sequence. You need to generate a declarative sentence that, when combined with the content of the first image, will enable a search engine to accurately retrieve the second image.

        Output Example:
        
        ``should add more people in a bright room.''

        Requirement:
        
        1. The output sentence should describe the modifications needed to change the first image into the second image.
        
        2. The subject of the output sentence should be ``the first image'' and the predicate can be a series of verbs such as ``increase'', ``enlarge'', ``reduce'', ``show'', ``zoom'', etc. However, the output must follow the given output example, and the subject must be omitted.
        
        3. Only this sentence can be output, and no other characters are allowed.
        
        4. The declarative sentence must be output in English.
        
        5. The modifications described in the declarative sentence should focus more on the elements within the images (people, objects, colors, numbers, environments, etc.).

        \{Input Image Sequence\}

\end{tcolorbox}
\label{tab:prompt_3}
\end{table}

\begin{table}[!htb]
\caption{\small Prompt for Generating Final Relative Captions.}
\begin{tcolorbox}[
        colback=gray!10,
        colframe=black,
        width=\columnwidth,  
        arc=2mm, 
        auto outer arc,
        title={Prompt for generating final relative captions}]		
        \footnotesize  
        
        Given several declarative sentences without subjects but with similar meanings, you need to generate a declarative sentence in the same format. 

    However, the generated declarative sentence should be able to summarize all the given declarative sentences.
    
    I will start with ``Given declarative sentences:'' to provide you with the declarative sentences.
    
    When generating, strictly follow the given example, and do not generate any other characters. The generated summary declarative sentence must be simple enough and generated in one line.

    Input example:
    
    [``shows a person standing alone in a room with patterned wallpaper and no visible candles.'',
    
    ``shows a single person standing in front of a patterned wall.'',
    
    ``shows a lone individual standing indoors with a different background and lighting.'',
    
    ``shows a different woman in a dimly lit area with fewer visible light sources.'',
    
    ``shows a person alone in dim lighting.'']

    Generated example:
    
    ``shows a person standing alone.''

    \{Given declarative sentences\}

\end{tcolorbox}
\label{tab:prompt_4}
\end{table}

\section{Discussions}

\subsection{Categorization of The Noise Ratio in Existing Benchmarks}
\label{categorization}
We randomly sampled 200 query pairs from the validation sets of CIRCO, CIRR, and FashionIQ, and manually reviewed the number of content-inconsistent and reference-irrelevant queries in each dataset's corresponding 200 pairs. Notes: (1) To ensure a fair comparison, we adopted a relaxed content-consistency criterion (requiring only core elements to match, acknowledging inherent subjectivity) rather than the stricter ZeroSight definition (which demands exact matches, e.g., from the same video/person). Three annotators independently labeled each query, and results reflect their average scores. (2) For CIRCO, queries targeting multiple images were counted as separate cases. (3) For FashionIQ, we sampled evenly across the Shirt, Dress, and Toptee subcategories.
%

The results are shown in Table~\ref{tab:cate_noisy}, which reveals that FashionIQ exhibits the highest proportion of both noise types, with content inconsistency reaching $58.0\%$ and reference irrelevance at $17.0\%$. This is largely attributable to the practical impossibility of maintaining identical models across images in this domain. Furthermore, CIRCO and CIRR also show significant noise levels, with the combined proportion of these issues approaching one-third of queries. These findings demonstrate that content-inconsistent and reference-irrelevant noise is prevalent in existing benchmarks. While this initial analysis is based on a sample of 200 queries per dataset, we plan to extend this verification to the full datasets and will open-source the identified noisy case IDs to facilitate future dataset refinement efforts.

\begin{table}[!t]
\caption{\small Categorization of The Noise Ratio in CIRCO, CIRR and FashionIQ}
\centering
\footnotesize
\begin{tabular}{@{}cccccccc@{}}
\toprule
\textbf{Benchmarks}     & \textbf{Content-Inconsistent} & \textbf{Reference-Irrelevant} \\ \midrule
CIRCO     & 26.66\%             & 6\%                         \\
CIRR      & 32.33\%             & 8\%                        \\
FashionIQ    & 58.00\%             & 17\%                \\ \bottomrule
\end{tabular}
\label{tab:cate_noisy}
\end{table}

\begin{figure}[!t]
\centering
\vspace{0.2cm}
\includegraphics[width=3.5in]{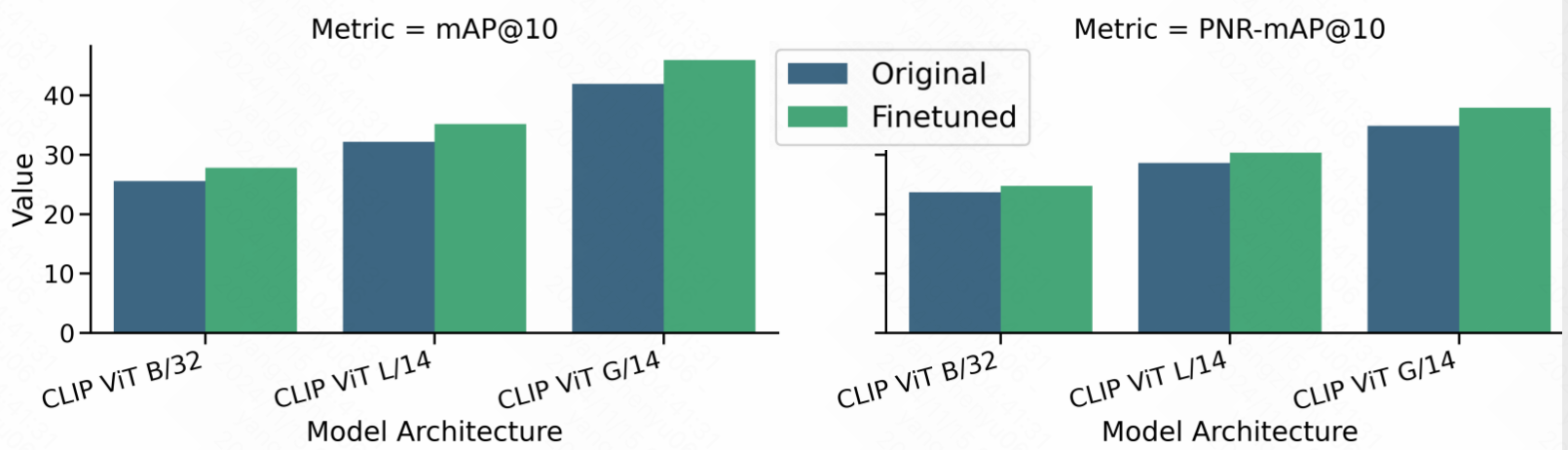}
\caption{\small Performance difference in ZS-CIR before and after training on ZeroSight image data. The results demonstrate that pre-training on the same dataset inflates ZS-CIR performance, with larger CLIP models exhibiting more significant improvements.}
\label{fig:impact}
    \vspace{-5mm}
\end{figure}

\subsection{Negative Impact of Pre-trained CLIP on ZS-CIR}
\label{negative_impact}
Given that the image pairs in the current ZS-CIR datasets are derived from image datasets that may have been pre-trained by CLIP, resulting in a spurious zero-shot scenario, it is crucial to demonstrate that this spurious zero-shot scenario can adversely affect ZS-CIR performance. We design an experiment to compare the performance of CLIP on the ZS-CIR task before and after training on ZeroSight image data. Since ZeroSight image data originates from video datasets post-2022, performing ZS-CIR directly on it will not introduce the influence of pre-training. Subsequently, we conduct contrastive learning training on the ZeroSight image data using image-caption pairs (captions generated by BLIP-2~\cite{li2023blip}). 
The performance improvement in ZS-CIR following this training indicates the impact of CLIP pre-training. We consistently employ LDRE~\cite{yang2024ldre} for the CIR experiments. The experimental results, presented in Figure \ref{fig:impact}, demonstrate that for all CLIP architectures, pre-training on a specific image dataset leads to inflated results when performing composed image retrieval on ZS-CIR datasets constructed from the same dataset. Furthermore, the larger the CLIP model, the more pronounced the inflation; for instance, CLIP ViT-G/14 exhibited a 8.68\% improvement.

\subsection{Application of ZeroSight and SC4CIR in Real-World}
\label{application}
As far as we know, in the medical field, when retrieving medical images, and in remote sensing image analysis tasks for urban planning and environmental monitoring, the target image obtained by modifying the reference image according to the relative caption must be very similar to the original reference image in the unmodified areas. This ensures the usability of the retrieval results. In these applications, the differences between images can be very subtle, yet the accuracy requirements for retrieval are very high. The practicality of our dataset in the real world lies in these applications. In correspondence with ZeroSight, \textit{SC4CIR} is also better suited for applications that demand extremely high retrieval accuracy. In these applications, errors in retrieval results can lead to very serious consequences, so the importance of retrieval accuracy far outweighs the real-time nature of retrieval.

\section{Conclusion}
\label{conclusion}
In this paper, we introduce ZeroSight, a novel benchmark for ZS-CIR that addresses the limitations of existing datasets by ensuring visual and semantic consistency through video-sourced data, ensuring a true zero-shot scenario by using data that has not been included in the pre-training of models like CLIP. Our multi-stage LLM-assisted pipeline generates high-quality queries with multiple positive and negative target images, providing a robust evaluation framework.
Additionally, we introduce a novel, training-free MLLM-driven ZS-CIR method called SC4CIR, designed to identify hard negative targets in retrieval tasks.
%
%
Experiments show that current CLIP-based CIR methods perform inflatedly on existing datasets, highlighting the need for ZeroSight. Our benchmark sets a new standard for both ZS-CIR and CIR research, fostering the development of more reliable and generalizable methods.

\bibliographystyle{IEEEtran}
\bibliography{references}

@inproceedings{vo2019composing,
  title={Composing text and image for image retrieval-an empirical odyssey},
  author={Vo, Nam and Jiang, Lu and Sun, Chen and Murphy, Kevin and Li, Li-Jia and Fei-Fei, Li and Hays, James},
  booktitle={Proceedings of the IEEE/CVF conference on computer vision and pattern recognition},
  pages={6439--6448},
  year={2019}
}

@inproceedings{chen2020image,
  title={Image search with text feedback by visiolinguistic attention learning},
  author={Chen, Yanbei and Gong, Shaogang and Bazzani, Loris},
  booktitle={Proceedings of the IEEE/CVF Conference on Computer Vision and Pattern Recognition},
  pages={3001--3011},
  year={2020}
}

@inproceedings{gu2021image,
  title={Image search with text feedback by deep hierarchical attention mutual information maximization},
  author={Gu, Chunbin and Bu, Jiajun and Zhang, Zhen and Yu, Zhi and Ma, Dongfang and Wang, Wei},
  booktitle={Proceedings of the 29th ACM International Conference on Multimedia},
  pages={4600--4609},
  year={2021}
}

@inproceedings{lee2021cosmo,
  title={Cosmo: Content-style modulation for image retrieval with text feedback},
  author={Lee, Seungmin and Kim, Dongwan and Han, Bohyung},
  booktitle={Proceedings of the IEEE/CVF Conference on Computer Vision and Pattern Recognition},
  pages={802--812},
  year={2021}
}

@inproceedings{kim2021dual,
  title={Dual compositional learning in interactive image retrieval},
  author={Kim, Jongseok and Yu, Youngjae and Kim, Hoeseong and Kim, Gunhee},
  booktitle={Proceedings of the AAAI Conference on Artificial Intelligence},
  volume={35},
  number={2},
  pages={1771--1779},
  year={2021}
}

@inproceedings{hou2020visual,
  title={Visual compositional learning for human-object interaction detection},
  author={Hou, Zhi and Peng, Xiaojiang and Qiao, Yu and Tao, Dacheng},
  booktitle={Computer Vision--ECCV 2020: 16th European Conference, Glasgow, UK, August 23--28, 2020, Proceedings, Part XV 16},
  pages={584--600},
  year={2020},
  organization={Springer}
}

@inproceedings{baldrati2022effective,
  title={Effective conditioned and composed image retrieval combining clip-based features},
  author={Baldrati, Alberto and Bertini, Marco and Uricchio, Tiberio and Del Bimbo, Alberto},
  booktitle={Proceedings of the IEEE/CVF Conference on Computer Vision and Pattern Recognition},
  pages={21466--21474},
  year={2022}
}

@inproceedings{wu2021fashion,
  title={Fashion iq: A new dataset towards retrieving images by natural language feedback},
  author={Wu, Hui and Gao, Yupeng and Guo, Xiaoxiao and Al-Halah, Ziad and Rennie, Steven and Grauman, Kristen and Feris, Rogerio},
  booktitle={Proceedings of the IEEE/CVF Conference on computer vision and pattern recognition},
  pages={11307--11317},
  year={2021}
}

@inproceedings{saito2023pic2word,
  title={Pic2word: Mapping pictures to words for zero-shot composed image retrieval},
  author={Saito, Kuniaki and Sohn, Kihyuk and Zhang, Xiang and Li, Chun-Liang and Lee, Chen-Yu and Saenko, Kate and Pfister, Tomas},
  booktitle={Proceedings of the IEEE/CVF Conference on Computer Vision and Pattern Recognition},
  pages={19305--19314},
  year={2023}
}

@article{baldrati2023zero,
  title={Zero-Shot Composed Image Retrieval with Textual Inversion},
  author={Baldrati, Alberto and Agnolucci, Lorenzo and Bertini, Marco and Del Bimbo, Alberto},
  journal={arXiv preprint arXiv:2303.15247},
  year={2023}
}

@article{tang2023context,
  title={Context-I2W: Mapping Images to Context-dependent Words for Accurate Zero-Shot Composed Image Retrieval},
  author={Tang, Yuanmin and Yu, Jing and Gai, Keke and Jiamin, Zhuang and Xiong, Gang and Hu, Yue and Wu, Qi},
  journal={arXiv preprint arXiv:2309.16137},
  year={2023}
}

@inproceedings{radford2021learning,
  title={Learning transferable visual models from natural language supervision},
  author={Radford, Alec and Kim, Jong Wook and Hallacy, Chris and Ramesh, Aditya and Goh, Gabriel and Agarwal, Sandhini and Sastry, Girish and Askell, Amanda and Mishkin, Pamela and Clark, Jack and others},
  booktitle={International conference on machine learning},
  pages={8748--8763},
  year={2021},
  organization={PMLR}
}

@inproceedings{yang2024ldre,
  title={LDRE: LLM-based Divergent Reasoning and Ensemble for Zero-Shot Composed Image Retrieval},
  author={Yang, Zhenyu and Xue, Dizhan and Qian, Shengsheng and Dong, Weiming and Xu, Changsheng},
  booktitle={Proceedings of the 47th International ACM SIGIR Conference on Research and Development in Information Retrieval},
  pages={80--90},
  year={2024}
}

@article{devlin2018bert,
  title={Bert: Pre-training of deep bidirectional transformers for language understanding},
  author={Devlin, Jacob and Chang, Ming-Wei and Lee, Kenton and Toutanova, Kristina},
  journal={arXiv preprint arXiv:1810.04805},
  year={2018}
}

@inproceedings{chen2020uniter,
  title={Uniter: Universal image-text representation learning},
  author={Chen, Yen-Chun and Li, Linjie and Yu, Licheng and El Kholy, Ahmed and Ahmed, Faisal and Gan, Zhe and Cheng, Yu and Liu, Jingjing},
  booktitle={European conference on computer vision},
  pages={104--120},
  year={2020},
  organization={Springer}
}

@article{li2019visualbert,
  title={Visualbert: A simple and performant baseline for vision and language},
  author={Li, Liunian Harold and Yatskar, Mark and Yin, Da and Hsieh, Cho-Jui and Chang, Kai-Wei},
  journal={arXiv preprint arXiv:1908.03557},
  year={2019}
}

@inproceedings{li2020oscar,
  title={Oscar: Object-semantics aligned pre-training for vision-language tasks},
  author={Li, Xiujun and Yin, Xi and Li, Chunyuan and Zhang, Pengchuan and Hu, Xiaowei and Zhang, Lei and Wang, Lijuan and Hu, Houdong and Dong, Li and Wei, Furu and others},
  booktitle={Computer Vision--ECCV 2020: 16th European Conference, Glasgow, UK, August 23--28, 2020, Proceedings, Part XXX 16},
  pages={121--137},
  year={2020},
  organization={Springer}
}

@article{lu2019vilbert,
  title={Vilbert: Pretraining task-agnostic visiolinguistic representations for vision-and-language tasks},
  author={Lu, Jiasen and Batra, Dhruv and Parikh, Devi and Lee, Stefan},
  journal={Advances in neural information processing systems},
  volume={32},
  year={2019}
}

@article{tan2019lxmert,
  title={Lxmert: Learning cross-modality encoder representations from transformers},
  author={Tan, Hao and Bansal, Mohit},
  journal={arXiv preprint arXiv:1908.07490},
  year={2019}
}

@article{vaswani2017attention,
  title={Attention is all you need},
  author={Vaswani, Ashish and Shazeer, Noam and Parmar, Niki and Uszkoreit, Jakob and Jones, Llion and Gomez, Aidan N and Kaiser, {\L}ukasz and Polosukhin, Illia},
  journal={Advances in neural information processing systems},
  volume={30},
  year={2017}
}

@inproceedings{baldrati2022conditioned,
  title={Conditioned and composed image retrieval combining and partially fine-tuning clip-based features},
  author={Baldrati, Alberto and Bertini, Marco and Uricchio, Tiberio and Del Bimbo, Alberto},
  booktitle={Proceedings of the IEEE/CVF Conference on Computer Vision and Pattern Recognition},
  pages={4959--4968},
  year={2022}
}

@inproceedings{han2023fame,
  title={FAME-ViL: Multi-Tasking Vision-Language Model for Heterogeneous Fashion Tasks},
  author={Han, Xiao and Zhu, Xiatian and Yu, Licheng and Zhang, Li and Song, Yi-Zhe and Xiang, Tao},
  booktitle={Proceedings of the IEEE/CVF Conference on Computer Vision and Pattern Recognition},
  pages={2669--2680},
  year={2023}
}

@article{karthik2023vision,
  title={Vision-by-Language for Training-Free Compositional Image Retrieval},
  author={Karthik, Shyamgopal and Roth, Karsten and Mancini, Massimiliano and Akata, Zeynep},
  journal={arXiv preprint arXiv:2310.09291},
  year={2023}
}

@inproceedings{vaze2023genecis,
  title={Genecis: A benchmark for general conditional image similarity},
  author={Vaze, Sagar and Carion, Nicolas and Misra, Ishan},
  booktitle={Proceedings of the IEEE/CVF Conference on Computer Vision and Pattern Recognition},
  pages={6862--6872},
  year={2023}
}

@inproceedings{liu2021image,
  title={Image retrieval on real-life images with pre-trained vision-and-language models},
  author={Liu, Zheyuan and Rodriguez-Opazo, Cristian and Teney, Damien and Gould, Stephen},
  booktitle={Proceedings of the IEEE/CVF International Conference on Computer Vision},
  pages={2125--2134},
  year={2021}
}

@inproceedings{lin2014microsoft,
  title={Microsoft coco: Common objects in context},
  author={Lin, Tsung-Yi and Maire, Michael and Belongie, Serge and Hays, James and Perona, Pietro and Ramanan, Deva and Doll{\'a}r, Piotr and Zitnick, C Lawrence},
  booktitle={Computer Vision--ECCV 2014: 13th European Conference, Zurich, Switzerland, September 6-12, 2014, Proceedings, Part V 13},
  pages={740--755},
  year={2014},
  organization={Springer}
}

@inproceedings{suhr2017corpus,
  title={A corpus of natural language for visual reasoning},
  author={Suhr, Alane and Lewis, Mike and Yeh, James and Artzi, Yoav},
  booktitle={Proceedings of the 55th Annual Meeting of the Association for Computational Linguistics (Volume 2: Short Papers)},
  pages={217--223},
  year={2017}
}

@inproceedings{deng2009imagenet,
  title={Imagenet: A large-scale hierarchical image database},
  author={Deng, Jia and Dong, Wei and Socher, Richard and Li, Li-Jia and Li, Kai and Fei-Fei, Li},
  booktitle={2009 IEEE conference on computer vision and pattern recognition},
  pages={248--255},
  year={2009},
  organization={Ieee}
}

@article{schuhmann2022laion,
  title={Laion-5b: An open large-scale dataset for training next generation image-text models},
  author={Schuhmann, Christoph and Beaumont, Romain and Vencu, Richard and Gordon, Cade and Wightman, Ross and Cherti, Mehdi and Coombes, Theo and Katta, Aarush and Mullis, Clayton and Wortsman, Mitchell and others},
  journal={Advances in Neural Information Processing Systems},
  volume={35},
  pages={25278--25294},
  year={2022}
}

@article{gadre2024datacomp,
  title={Datacomp: In search of the next generation of multimodal datasets},
  author={Gadre, Samir Yitzhak and Ilharco, Gabriel and Fang, Alex and Hayase, Jonathan and Smyrnis, Georgios and Nguyen, Thao and Marten, Ryan and Wortsman, Mitchell and Ghosh, Dhruba and Zhang, Jieyu and others},
  journal={Advances in Neural Information Processing Systems},
  volume={36},
  year={2024}
}

@inproceedings{srinivasan2021wit,
  title={Wit: Wikipedia-based image text dataset for multimodal multilingual machine learning},
  author={Srinivasan, Krishna and Raman, Karthik and Chen, Jiecao and Bendersky, Michael and Najork, Marc},
  booktitle={Proceedings of the 44th international ACM SIGIR conference on research and development in information retrieval},
  pages={2443--2449},
  year={2021}
}

@article{chen2022pali,
  title={Pali: A jointly-scaled multilingual language-image model},
  author={Chen, Xi and Wang, Xiao and Changpinyo, Soravit and Piergiovanni, AJ and Padlewski, Piotr and Salz, Daniel and Goodman, Sebastian and Grycner, Adam and Mustafa, Basil and Beyer, Lucas and others},
  journal={arXiv preprint arXiv:2209.06794},
  year={2022}
}

@article{fang2023data,
  title={Data filtering networks},
  author={Fang, Alex and Jose, Albin Madappally and Jain, Amit and Schmidt, Ludwig and Toshev, Alexander and Shankar, Vaishaal},
  journal={arXiv preprint arXiv:2309.17425},
  year={2023}
}

@inproceedings{yang2024semantic,
  title={Semantic Editing Increment Benefits Zero-Shot Composed Image Retrieval},
  author={Yang, Zhenyu and Qian, Shengsheng and Xue, Dizhan and Wu, Jiahong and Yang, Fan and Dong, Weiming and Xu, Changsheng},
  booktitle={Proceedings of the 32nd ACM International Conference on Multimedia},
  pages={1245--1254},
  year={2024}
}

@inproceedings{cohen2022my,
  title={“This is my unicorn, Fluffy”: Personalizing frozen vision-language representations},
  author={Cohen, Niv and Gal, Rinon and Meirom, Eli A and Chechik, Gal and Atzmon, Yuval},
  booktitle={European Conference on Computer Vision},
  pages={558--577},
  year={2022},
  organization={Springer}
}

@article{gu2023language,
  title={Language-only Efficient Training of Zero-shot Composed Image Retrieval},
  author={Gu, Geonmo and Chun, Sanghyuk and Kim, Wonjae and Kang, Yoohoon and Yun, Sangdoo},
  journal={arXiv preprint arXiv:2312.01998},
  year={2023}
}

@inproceedings{zhang2021heterogeneous,
  title={Heterogeneous feature fusion and cross-modal alignment for composed image retrieval},
  author={Zhang, Gangjian and Wei, Shikui and Pang, Huaxin and Zhao, Yao},
  booktitle={Proceedings of the 29th ACM International Conference on Multimedia},
  pages={5353--5362},
  year={2021}
}

@inproceedings{wen2023target,
  title={Target-guided composed image retrieval},
  author={Wen, Haokun and Zhang, Xian and Song, Xuemeng and Wei, Yinwei and Nie, Liqiang},
  booktitle={Proceedings of the 31st ACM International Conference on Multimedia},
  pages={915--923},
  year={2023}
}

@inproceedings{li2023blip,
  title={Blip-2: Bootstrapping language-image pre-training with frozen image encoders and large language models},
  author={Li, Junnan and Li, Dongxu and Savarese, Silvio and Hoi, Steven},
  booktitle={International conference on machine learning},
  pages={19730--19742},
  year={2023},
  organization={PMLR}
}

@article{achiam2023gpt,
  title={Gpt-4 technical report},
  author={Achiam, Josh and Adler, Steven and Agarwal, Sandhini and Ahmad, Lama and Akkaya, Ilge and Aleman, Florencia Leoni and Almeida, Diogo and Altenschmidt, Janko and Altman, Sam and Anadkat, Shyamal and others},
  journal={arXiv preprint arXiv:2303.08774},
  year={2023}
}

@article{dosovitskiy2020image,
  title={An image is worth 16x16 words: Transformers for image recognition at scale},
  author={Dosovitskiy, Alexey},
  journal={arXiv preprint arXiv:2010.11929},
  year={2020}
}

@inproceedings{zellers2022merlot,
  title={Merlot reserve: Neural script knowledge through vision and language and sound},
  author={Zellers, Rowan and Lu, Jiasen and Lu, Ximing and Yu, Youngjae and Zhao, Yanpeng and Salehi, Mohammadreza and Kusupati, Aditya and Hessel, Jack and Farhadi, Ali and Choi, Yejin},
  booktitle={Proceedings of the IEEE/CVF Conference on Computer Vision and Pattern Recognition},
  pages={16375--16387},
  year={2022}
}

@article{karthik2024visionbylanguage,
  title={Vision-by-Language for Training-Free Compositional Image Retrieval},
  author={Shyamgopal Karthik and Karsten Roth and Massimiliano Mancini and Zeynep Akata},
  journal={International Conference on Learning Representations (ICLR)},
  year={2024}
}

@inproceedings{eccv2022_palavra_cohen,
 author = {Cohen, Niv and Gal, Rinon and Meirom, Eli A. and Chechik, Gal and Atzmon, Yuval},
 booktitle = {European Conference on Computer Vision (ECCV) },
 title = {"This is my unicorn, Fluffy": Personalizing frozen vision-language representations},
 year = {2022}
}

@article{agnolucci2024isearle,
  title={iSEARLE: Improving Textual Inversion for Zero-Shot Composed Image Retrieval}, 
  author={Agnolucci, Lorenzo and Baldrati, Alberto and Bertini, Marco and Del Bimbo, Alberto},
  journal={arXiv preprint arXiv:2405.02951},
  year={2024}
}

@inproceedings{gu2024lincir,
    title={Language-only Training of Zero-shot Composed Image Retrieval},
    author={Gu, Geonmo and Chun, Sanghyuk and Kim, Wonjae and and Kang, Yoohoon and Yun, Sangdoo},
    year={2024},
    booktitle={Conference on Computer Vision and Pattern Recognition (CVPR)}
}

@article{delmas2022artemis,
  title={Artemis: Attention-based retrieval with text-explicit matching and implicit similarity},
  author={Delmas, Ginger and de Rezende, Rafael Sampaio and Csurka, Gabriela and Larlus, Diane},
  journal={arXiv preprint arXiv:2203.08101},
  year={2022}
}

@article{zhu2023amc,
  title={AMC: Adaptive Multi-expert Collaborative Network for Text-guided Image Retrieval},
  author={Zhu, Hongguang and Wei, Yunchao and Zhao, Yao and Zhang, Chunjie and Huang, Shujuan},
  journal={ACM Transactions on Multimedia Computing, Communications and Applications},
  year={2023}
}

@article{dodds2020modality,
  title={Modality-Agnostic Attention Fusion for visual search with text feedback},
  author={Dodds, Eric and Culpepper, Jack and Herdade, Simao and Zhang, Yang and Boakye, Kofi},
  journal={arXiv preprint arXiv:2007.00145},
  year={2020}
}

@inproceedings{huang2023dynamic,
  title={Dynamic Weighted Combiner for Mixed-Modal Image Retrieval},
  author={Huang, Fuxiang and Zhang, Lei and Fu, Xiaowei and Song, Suqi},
  booktitle={Association for the Advance of Artificial Intelligence (AAAI)}, 
  year={2024}
}

@article{baldrati2023composed,
  title={Composed Image Retrieval using Contrastive Learning and Task-oriented CLIP-based Features},
  author={Baldrati, Alberto and Bertini, Marco and Uricchio, Tiberio and Bimbo, Alberto Del},
  journal={ACM Transactions on Multimedia Computing, Communications and Applications},
  publisher={ACM New York, NY}
}

@inproceedings{xu2024sentence,
  title={Sentence-level prompts benefit composed image retrieval},
  author={Xu, Xinxing and Liu, Yong and Khan, Salman and Khan, Fahad and Zuo, Wangmeng and Goh, Rick Siow Mong and Feng, Chun-Mei and others},
  booktitle={The Twelfth International Conference on Learning Representations},
  year={2024}
}

@article{liu2023zero,
  title={Zero-shot composed text-image retrieval},
  author={Liu, Yikun and Yao, Jiangchao and Zhang, Ya and Wang, Yanfeng and Xie, Weidi},
  journal={arXiv preprint arXiv:2306.07272},
  year={2023}
}

@inproceedings{sun2025leveraging,
  title={Leveraging large vision-language model as user intent-aware encoder for composed image retrieval},
  author={Sun, Zelong and Jing, Dong and Yang, Guoxing and Fei, Nanyi and Lu, Zhiwu},
  booktitle={Proceedings of the AAAI Conference on Artificial Intelligence},
  volume={39},
  number={7},
  pages={7149--7157},
  year={2025}
}

@article{tianccin,
  title={CCIN: Compositional Conflict Identification and Neutralization for Composed Image Retrieval},
  author={Tian, Likai and Zhao, Jian and Hu, Zechao and Yang, Zhengwei and Li, Hao and Jin, Lei and Wang, Zheng and Li, Xuelong}
}

@inproceedings{bao2025mllm,
  title={MLLM-I2W: Harnessing Multimodal Large Language Model for Zero-Shot Composed Image Retrieval},
  author={Bao, Tong and Liu, Che and Xu, Derong and Zheng, Zhi and Xu, Tong},
  booktitle={Proceedings of the 31st International Conference on Computational Linguistics},
  pages={1839--1849},
  year={2025}
}

@article{wanggenerative,
  title={Generative Zero-Shot Composed Image Retrieval},
  author={Wang, Lan and Ao, Wei and Boddeti, Vishnu Naresh and Lim, Ser-Nam}
}

@inproceedings{ventura2024covr,
  title={CoVR: Learning composed video retrieval from web video captions},
  author={Ventura, Lucas and Yang, Antoine and Schmid, Cordelia and Varol, G{\"u}l},
  booktitle={Proceedings of the AAAI Conference on Artificial Intelligence},
  volume={38},
  number={6},
  pages={5270--5279},
  year={2024}
}

@inproceedings{liu2024bi,
  title={Bi-directional training for composed image retrieval via text prompt learning},
  author={Liu, Zheyuan and Sun, Weixuan and Hong, Yicong and Teney, Damien and Gould, Stephen},
  booktitle={Proceedings of the IEEE/CVF Winter Conference on Applications of Computer Vision},
  pages={5753--5762},
  year={2024}
}

@article{Levy_Ben-Ari_Darshan_Lischinski_2024,
 title={Data Roaming and Quality Assessment for Composed Image Retrieval},
  volume={38}, url={https://ojs.aaai.org/index.php/AAAI/article/view/28081},
  DOI={10.1609/aaai.v38i4.28081},
  abstractNote={The task of Composed Image Retrieval (CoIR) involves queries that combine image and text modalities, allowing users to express their intent more effectively. However, current CoIR datasets are orders of magnitude smaller compared to other vision and language (V&amp;L) datasets. Additionally, some of these datasets have noticeable issues, such as queries containing redundant modalities. To address these shortcomings, we introduce the Large Scale Composed Image Retrieval (LaSCo) dataset, a new CoIR dataset which is ten times larger than existing ones. Pre-training on our LaSCo, shows a noteworthy improvement in performance, even in zero-shot. Furthermore, we propose a new approach for analyzing CoIR datasets and methods, which detects modality redundancy or necessity, in queries.
    We also introduce a new CoIR baseline, the Cross-Attention driven Shift Encoder (CASE). This baseline allows for early fusion of modalities using a cross-attention module and employs an additional auxiliary task during training. Our experiments demonstrate that this new baseline outperforms the current state-of-the-art methods on established benchmarks like FashionIQ and CIRR.},
  number={4},
  journal={Proceedings of the AAAI Conference on Artificial Intelligence},
  author={Levy, Matan and Ben-Ari, Rami and Darshan, Nir and Lischinski, Dani},
  year={2024},
  month={Mar.},
  pages={2991-2999}
}

@article{team2023gemini,
  title={Gemini: a family of highly capable multimodal models},
  author={Team, Gemini and Anil, Rohan and Borgeaud, Sebastian and Alayrac, Jean-Baptiste and Yu, Jiahui and Soricut, Radu and Schalkwyk, Johan and Dai, Andrew M and Hauth, Anja and Millican, Katie and others},
  journal={arXiv preprint arXiv:2312.11805},
  year={2023}
}

@article{touvron2023llama,
  title={LLaMA: Open and Efficient Foundation Language Models},
  author={Touvron, Hugo and Lavril, Thibaut and Izacard, Gautier and Martinet, Xavier and Lachaux, Marie-Anne and Lacroix, Timoth{\'e}e and Rozi{\`e}re, Baptiste and Goyal, Naman and Hambro, Eric and Azhar, Faisal and Rodriguez, Aurelien and Joulin, Armand and Grave, Edouard and Lample, Guillaume},
  journal={arXiv preprint arXiv:2302.13971},
  year={2023}
}

@inproceedings{tang2025reason,
  title={Reason-before-retrieve: One-stage reflective chain-of-thoughts for training-free zero-shot composed image retrieval},
  author={Tang, Yuanmin and Zhang, Jue and Qin, Xiaoting and Yu, Jing and Gou, Gaopeng and Xiong, Gang and Lin, Qingwei and Rajmohan, Saravan and Zhang, Dongmei and Wu, Qi},
  booktitle={Proceedings of the Computer Vision and Pattern Recognition Conference},
  pages={14400--14410},
  year={2025}
}

@article{wen2023self,
  title={Self-training boosted multi-factor matching network for composed image retrieval},
  author={Wen, Haokun and Song, Xuemeng and Yin, Jianhua and Wu, Jianlong and Guan, Weili and Nie, Liqiang},
  journal={IEEE Transactions on Pattern Analysis and Machine Intelligence},
  volume={46},
  number={5},
  pages={3665--3678},
  year={2023},
  publisher={IEEE}
}

@article{takahashi2014mixture,
  title={Mixture of subspaces image representation and compact coding for large-scale image retrieval},
  author={Takahashi, Takashi and Kurita, Takio},
  journal={IEEE transactions on pattern analysis and machine intelligence},
  volume={37},
  number={7},
  pages={1469--1479},
  year={2014},
  publisher={IEEE}
}

@article{gao2020fashion,
  title={Fashion retrieval via graph reasoning networks on a similarity pyramid},
  author={Gao, Yiming and Kuang, Zhanghui and Li, Guanbin and Luo, Ping and Chen, Yimin and Lin, Liang and Zhang, Wayne},
  journal={IEEE Transactions on Pattern Analysis and Machine Intelligence},
  volume={45},
  number={6},
  pages={7019--7034},
  year={2020},
  publisher={IEEE}
}

@article{tan2025dynamic,
  title={Dynamic Bit-Wise Semantic Transformer Hashing for Multi-Modal Retrieval},
  author={Tan, Wentao and Li, Fengling and Zhu, Lei and Guan, Weili and Li, Jingjing and Cheng, Zhiyong and Shen, Heng Tao},
  journal={IEEE Transactions on Pattern Analysis and Machine Intelligence},
  year={2025},
  publisher={IEEE}
}

@article{cao2025multilingual,
  title={Multilingual Text-to-Image Person Retrieval via Bidirectional Relation Reasoning and Aligning},
  author={Cao, Min and Zhou, Xinyu and Jiang, Ding and Du, Bo and Ye, Mang and Zhang, Min},
  journal={IEEE Transactions on Pattern Analysis and Machine Intelligence},
  year={2025},
  publisher={IEEE}
}

@article{chen2025prvr,
  title={Prvr: Partially relevant video retrieval},
  author={Chen, Xianke and Liu, Daizong and Yang, Xun and Li, Xirong and Dong, Jianfeng and Wang, Meng and Wang, Xun},
  journal={IEEE Transactions on Pattern Analysis and Machine Intelligence},
  year={2025},
  publisher={IEEE}
}

@article{agnolucci2025isearle,
  title={isearle: Improving textual inversion for zero-shot composed image retrieval},
  author={Agnolucci, Lorenzo and Baldrati, Alberto and Del Bimbo, Alberto and Bertini, Marco},
  journal={IEEE Transactions on Pattern Analysis and Machine Intelligence},
  year={2025},
  publisher={IEEE}
}

@article{zhang2025active,
  title={Active supervised cross-modal retrieval},
  author={Zhang, Huaiwen and Yang, Yang and Qi, Fan and Qian, Shengsheng and Xu, Changsheng},
  journal={IEEE Transactions on Pattern Analysis and Machine Intelligence},
  year={2025},
  publisher={IEEE}
}

@article{ramzi2025optimization,
  title={Optimization of rank losses for image retrieval},
  author={Ramzi, Elias and Audebert, Nicolas and Rambour, Cl{\'e}ment and Araujo, Andr{\'e} and Bitot, Xavier and Thome, Nicolas},
  journal={IEEE Transactions on Pattern Analysis and Machine Intelligence},
  year={2025},
  publisher={IEEE}
}

@article{li2025attack,
  title={Attack as Defense: Proactive Adversarial Multi-Modal Learning to Evade Retrieval},
  author={Li, Fengling and Wang, Tianshi and Zhu, Lei and Li, Jingjing and Shen, Heng Tao},
  journal={IEEE Transactions on Pattern Analysis and Machine Intelligence},
  year={2025},
  publisher={IEEE}
}

@article{wang2023towards,
  title={Towards codebook-free deep probabilistic quantization for image retrieval},
  author={Wang, Min and Zhou, Wengang and Yao, Xin and Tian, Qi and Li, Houqiang},
  journal={IEEE Transactions on Pattern Analysis and Machine Intelligence},
  volume={46},
  number={1},
  pages={626--640},
  year={2023},
  publisher={IEEE}
}

@inproceedings{yangsvbench,
  title={SVBench: A Benchmark with Temporal Multi-Turn Dialogues for Streaming Video Understanding},
  author={Yang, Zhenyu and Hu, Yuhang and Du, Zemin and Xue, Dizhan and Qian, Shengsheng and Wu, Jiahong and Yang, Fan and Dong, Weiming and Xu, Changsheng},
  booktitle={The Thirteenth International Conference on Learning Representations}
}

@article{yang2026livestar,
  title={Livestar: Live streaming assistant for real-world online video understanding},
  author={Yang, Zhenyu and Zhang, Kairui and Hu, Yuhang and Wang, Bing and Qian, Shengsheng and Wen, Bin and Yang, Fan and Gao, Tingting and Dong, Weiming and Xu, Changsheng},
  journal={Advances in Neural Information Processing Systems},
  volume={38},
  pages={31266--31304},
  year={2026}
}

@inproceedings{zhang2026querystream,
  title={Querystream: Advancing streaming video understanding with query-aware pruning and proactive response},
  author={Zhang, Kairui and Yang, Zhenyu and Wang, Bing and Qian, Shengsheng and Xu, Changsheng},
  booktitle={The Fourteenth International Conference on Learning Representations},
  year={2026}
}

 




\vfill

\end{document}